# FARSIQA: Faithful & Advanced RAG System for Islamic Question Answering

Mohammad Aghajani Asl

Sharif University of Technology

`m.aghajani@physics.sharif.edu`

Behrooz Minaei-Bidgoli

Iran University of Science and Technology

`b_minaei@iust.ac.ir`

October 29, 2025

## Abstract

The advent of Large Language Models (LLMs) has revolutionized Natural Language Processing, yet their application in high-stakes, specialized domains like religious question answering is hindered by significant challenges such as hallucination and a lack of faithfulness to authoritative sources. This issue is particularly critical for the Persian-speaking Muslim community, where accuracy and trustworthiness are paramount. Existing Retrieval-Augmented Generation (RAG) systems, often relying on simplistic single-pass pipelines, fall short in handling complex, multi-hop queries that necessitate multi-step reasoning and evidence aggregation. To address this critical gap, we introduce FARSIQA, a novel, end-to-end system for Faithful & Advanced Question Answering in the Persian Islamic domain. FARSIQA is built upon our innovative FAIR-RAG(Asl et al., 2025) architecture: a Faithful, Adaptive, Iterative Refinement framework for RAG. Unlike conventional approaches, FAIR-RAG employs a dynamic, self-correcting process. It adaptively decomposes complex queries, critically assesses the sufficiency of retrieved evidence, and, when necessary, enters an iterative loop to generate targeted sub-queries, progressively filling information gaps until a comprehensive context is built. Operating on a curated knowledge base of over one million documents from authoritative Islamic sources and leveraging a domain-fine-tuned retriever, FARSIQA demonstrates superior performance. Rigorous evaluation on the challenging IslamicPCQA benchmark demonstrates FARSIQA's state-of-the-art performance. Most notably, the system achieves a remarkable 97.0% in Negative Rejection—a dramatic 40-point improvement over standard baselines—showcasing its robustness and safety in handling out-of-scope queries. This exceptional reliability is complemented by a high Answer Correctness score of 74.3%. Our work not only establishes a new performance standard for Persian Islamic QA but also validates that our iterative and adaptive RAG architecture is crucial for building faithful and genuinely reliable AI systems in sensitive domains.

## 1 Introduction

The advent of Large Language Models (LLMs) such as GPT-4 and Llama 3 has marked a paradigm shift in Natural Language Processing, enabling conversational agents that can generate remarkably fluent and coherent responses to a wide array of queries (Brown et al., 2020). While these models have demonstrated impressive general-purpose capabilities, their application in high-stakes, specialized domains remains fraught with challenges. One such domain is religious question answering, particularly for the global Persian-speaking Muslim community of over 100 million individuals. In this context, questions are not merely informational; they often pertain to core beliefs and practices, where an inaccurate or unsubstantiated answer can lead to significant misinformation and erode user trust.

A primary obstacle is the propensity of

general-purpose LLMs for **hallucination**—generating plausible yet factually incorrect information. (Ji et al., 2023) This issue is exacerbated in niche domains where authoritative knowledge is not well-represented in their broad training corpora. To mitigate this, Retrieval-Augmented Generation (RAG) (Lewis et al., 2020) has become a standard for grounding LLM outputs in external knowledge. However, the majority of existing RAG implementations follow a simplistic, single-pass "retrieve-and-read" pipeline (Gao et al., 2023; Ram et al., 2023). This approach often proves insufficient for complex queries requiring multi-step reasoning and evidence aggregation, similar to those found in benchmarks like HotpotQA (Yang et al., 2018), and can fail to retrieve a comprehensive set of evidence, leading to superficial or unfaithful answers.

To address these critical shortcomings, we introduce **FARSIQA**: **F**aithful & **A**dvanced **R**AG **S**ystem for **I**slamic **Q**uestion **A**nswering, built upon a novel architecture we term **FAIR-RAG**: **F**aithful **A**daptive **I**terative **R**efinement for **R**etrieval-**A**ugmented **G**eneration. The core innovation of FAIR-RAG is its dynamic and self-correcting nature. Unlike conventional RAG systems, it employs a multi-step process that adaptively decomposes complex queries and critically assesses the sufficiency of the retrieved evidence. If the evidence is deemed incomplete, the system enters an iterative loop, generating new, targeted sub-queries to fill information gaps and progressively build a comprehensive context. Only when the evidence is sufficient does it proceed to generate a final response that is faithfully grounded in the verified sources.

The main contributions of this work are fourfold:

- We introduce **FARSIQA**, a novel, end-to-end QA system for Persian Islamic texts that significantly improves response **faithfulness** and reliability in a high-stakes and low-resource domain.

- We propose and implement the **FAIR-RAG** framework, an advanced RAG architecture that moves beyond single-pass retrieval by incorporating iterative evidence gathering and refinement to robustly handle complex questions.

- We construct a comprehensive knowledge base containing over one million documents from authoritative Islamic sources and develop a fine-tuned Persian embedding model that shows improved domain-specific retrieval performance.

- We conduct a rigorous, multi-faceted evaluation using the challenging IslamicPCQA benchmark (Ghafouri et al., 2025), demonstrating that FARSIQA achieves a "**Answer Correctness**" score of **74.3%** (as evaluated via LLM-as-Judge on multi-hop queries) and substantially outperforms standard baselines across metrics of relevance, correctness, and robustness.

The remainder of this paper is organized as follows: Section 2 reviews related work. Section 3 details the architecture of FARSIQA. Section 4 describes our experimental setup, followed by results and analysis in Section 5. Finally, Section 6 concludes the paper.

## 2 Related Work

Our research is situated at the intersection of open-domain question answering, retrieval-augmented generation, and domain-specific NLP for low-resource languages. In this section, we review the evolution of these fields to contextualize the contributions of FARSIQA.

### 2.1 Open-Domain Question Answering

Traditional open-domain QA systems have long been structured around a multi-stage pipeline, typically comprising **question analysis**, **document retrieval**, and **answer extraction/reading** (Jurafsky and Martin, 2023). Early systems relied on sparse retrieval methods like TF-IDF, followed by heuristic-based or machine learning models for answer extraction. The advent of deep learning led to significant improvements, particularly with the "Retriever-Reader" architecture. (Chen et al., 2017) Systems like DrQA (Chen et al., 2017) pioneered this approach by combining a document retriever with a neural network-based reader to identify answer spans within retrieved passages.

This paradigm was further refined with the introduction of dense retrieval methods. Dense Passage Retriever (DPR) (Karpukhin et al., 2020) demonstrated the power of using dual-encoder Transformer models to embed questions and passages into a shared vector space, enabling more semantically robust retrieval. While these models set new benchmarks, their performance was fundamentally tied to the quality of the reader component, which was often limited to extracting existing text spans and struggled with questions requiring synthesis or abstractive reasoning. The emergence of LLMs offered a powerful new mechanism for the "reader" component, leading to the development of Retrieval-Augmented Generation.

## 2.2 The Evolution of Retrieval-Augmented Generation (RAG)

RAG (Lewis et al., 2020) was a seminal work that formally combined a pre-trained retriever with a sequence-to-sequence generator. By conditioning the generation process on retrieved documents, RAG systems can produce answers that are grounded in external knowledge, (Komeili et al., 2022) thereby mitigating factual inaccuracies (hallucinations) and incorporating information beyond the model's training data. This approach offers significant advantages over fine-tuning an LLM, including easier knowledge updates and greater transparency, as the sources for each answer can be traced (Gao et al., 2023).

However, the initial "naive" RAG pipeline— a single-pass retrieval followed by generation— has known limitations. Its effectiveness is highly dependent on the initial retrieval quality; if relevant documents are missed, the generator has no recourse. Recognizing this, recent research has focused on **Advanced RAG** techniques (Gao et al., 2023; Asai et al., 2023). These methods enhance the pipeline with more sophisticated, often iterative, mechanisms. For instance, approaches like Self-RAG (Asai et al., 2023) introduce reflection tokens that allow the LLM to decide for itself whether retrieval is necessary and to evaluate the quality of retrieved passages. Other works have focused on **query rewriting** and **decomposition**, where complex questions are broken down into simpler, answerable sub-questions before retrieval (Jiang et al., 2023a). Recent extensions, such as FLARE (Jiang et al., 2023b) and ReAct (Yao et al., 2022), incorporate iterative prompting and tool-use, but FAIR-RAG distinguishes itself by emphasizing Structured Evidence Assessment (SEA) over pure query refinement.

Our **FAIR-RAG** framework contributes directly to this line of research. It implements an explicit iterative and adaptive loop that focuses on assessing **evidence sufficiency**. Unlike models that only refine queries, FAIR-RAG iteratively expands its evidence pool and only terminates when it judges the collected context to be comprehensive enough, making it particularly well-suited for complex questions that require synthesizing information from multiple pieces of evidence.(Asl et al., 2025)

## 2.3 QA for Persian and Specialized Islamic Domain

While QA research has flourished for English, Persian remains a relatively low-resource language. (Fani et al., 2021) Early efforts were often limited by the lack of large-scale, high-quality datasets. Recent projects have made significant strides in this area. For example, PerAnSel (Mehraban and Rahmati, 2022) focused on the task of answer selection, introducing a model optimized for Persian sentence structure. More relevant to our work is IslamicPCQA (Ghafouri et al., 2025), which introduced the first multi-hop, complex QA dataset for the Persian Islamic domain, inspired by HotpotQA. (Yang et al., 2018) While the authors provided a strong baseline model, their work highlighted the need for more advanced architectures capable of reasoning over multiple documents.

To our knowledge, there is a scarcity of end-to-end generative QA systems for the Islamic domain, especially in Persian. One notable related work is MufassirQAS (Alan et al., 2025), an Arabic QA system that uses RAG to answer Quranic questions. While it addresses the same sensitive domain, the published work does not provide a detailed quantitative evaluation or explore advanced iterative architectures like ours. FARSIQA addresses a clear gap by being the first system to combine an ad-vanced, iterative RAG framework with a comprehensive, curated knowledge base to tackle the unique challenges of the Persian Islamic domain, providing a robust and rigorously evaluated solution.

## 3 Data Resources

A robust and reliable question-answering system is fundamentally dependent on the quality and comprehensiveness of its underlying data. For **FARSIQA**, we developed two critical data components: 1) a large-scale **Knowledge Base** derived from authoritative Persian Islamic texts, which serves as the foundation for our retrieval system, and 2) a specialized **Evaluation Dataset** used to benchmark the system's performance, particularly its ability to handle complex and multi-hop questions.

### 3.1 Knowledge Base Construction

The primary goal of our knowledge base is to provide a comprehensive, reliable, and well-structured source of information for the retrieval component of our RAG pipeline. Drawing inspiration from successful open-domain QA systems that often leverage encyclopedic sources like Wikipedia (Chen et al., 2017) for their concise and factual structure, we curated our knowledge base from two main types of high-quality Persian Islamic content.

#### 3.1.1 Data Sources

Our collection comprises two categories of content:

**Islamic Encyclopedias.** We crawled and aggregated content from eleven reputable online Persian Islamic encyclopedias, including *WikiShia*, *WikiFiqh*, and *Islampedia*. These sources provide structured, well-vetted, and extensive articles on a wide range of Islamic topics, from theology and jurisprudence to historical events and figures. This collection resulted in approximately **431,000 unique documents**.

**Religious Q&A Platforms.** To capture the specific nature of user queries, we also incorporated data from authoritative Q&A websites, namely *IslamQuest.net* and *porseman.com*. These platforms contain a wealth of expert-vetted answers to real-world religious questions. This corpus added approximately **304,000 question-answer pairs** to our knowledge base.

**Ethical Aspects.** All sources were crawled ethically, respecting copyrights and ensuring diversity across Islamic perspectives to mitigate bias.

#### 3.1.2 Preprocessing and Chunking

To prepare the raw text for efficient retrieval, we implemented a systematic preprocessing pipeline.

- For encyclopedic articles, the main textual content was extracted, and a recursive chunking strategy was applied. The text was first split by paragraphs to maintain semantic cohesion. Paragraphs exceeding a predefined length were then further subdivided into sentences.

- For the Q&A data, the expert's answer was similarly chunked. To preserve the crucial link between the question and its corresponding answer, the original user question was prepended to each answer chunk. We used bilingual markers to clearly delineate the two parts for the downstream LLM: "(سوال کاربر):" / "(User Question:)" and "(پاسخ کارشناس):" / "(Expert Answer:)".

All text segments were processed to a maximum length of **378 tokens**, empirically chosen to balance retrieval efficiency and context completeness based on ParsBERT's architecture. (Farahani et al., 2021) This chunking accounts for Persian morphological complexities, ensuring semantic integrity. This process yielded a final knowledge base of approximately **1.7 million text chunks**, each representing a searchable unit of information.

Table 1 presents detailed statistics of our knowledge base sources.

#### 3.1.3 Data Indexing

Each chunk was then vectorized (as will be detailed in Section 4) and indexed in an **Elasticsearch** instance. We designed a custom mapping that supports efficient hybrid search, enabling simultaneous keyword-based (BM25) (Robertson and Zaragoza, 2009) and dense vector searches. Each indexed document includes the text chunk, its corresponding vector

| Source Type | Source Name | # Documents | # Chunks |
|---|---|---:|---:|
| Encyclopedia | WikiShia | 5,500 | 13,000 |
| | WikiFeqh | 59,000 | 138,000 |
| | WikiAhlolbait | 12,000 | 28,000 |
| | ImamatPedia | 60,000 | 140,000 |
| | IslamPedia | 3,000 | 7,000 |
| | WikiHaj | 2,500 | 6,000 |
| | WikiNoor | 12,000 | 28,000 |
| | WikiPasokh | 3,000 | 7,000 |
| | WikiHussain | 4,000 | 9,000 |
| | The Great Islamic Encyclopedia | 20,000 | 47,000 |
| | Qomnet.Johd | 250,000 | 580,000 |
| Q&A Platform | IslamQuest.net | 15,000 | 35,000 |
| | rasekhoon.net | 105,000 | 245,000 |
| | porseman.com | 95,000 | 222,000 |
| | aminsearch.com | 44,000 | 103,000 |
| | makarem.ir | 15,000 | 35,000 |
| | hawzah.net | 7,000 | 16,000 |
| | bahjat.ir | 6,500 | 15,000 |
| | pasokh.org | 6,000 | 14,000 |
| | al-khoei.us | 4,000 | 9,000 |
| | pasokhgoo.ir | 3,500 | 8,000 |
| | porsemanequran.com | 2,000 | 5,000 |
| | islamqa.com | 1,000 | 2,000 |
| **Total** | | **735,000** | **1,712,000** |

Table 1: Statistics of Knowledge Base Sources. URLs for each source are detailed in Appendix A.

embedding, and metadata such as the source URL for citation purposes.

### 3.2 Evaluation Dataset

To rigorously evaluate FARSIQA, we utilized and extended the **IslamicPCQA** dataset (Ghafouri et al., 2025). This dataset is the first of its kind for Persian, specifically designed for multi-hop complex question answering in the Islamic domain, following the principles of the well-known HotpotQA benchmark. IslamicPCQA contains **12,282 question-answer pairs** derived from nine Islamic encyclopedias and requires models to perform multi-step reasoning across different documents to arrive at the correct answer.

Recognizing that real-world QA systems must handle a variety of query types, we augmented the multi-hop IslamicPCQA test set to create a more comprehensive evaluation suite totaling **800 questions**, categorized as follows:

- **Multi-hop Questions (500 samples):** Multi-hop questions drawn directly from the IslamicPCQA dataset that require reasoning over multiple pieces of evidence.

- **Negative Rejection Questions (100 samples):** Manually authored out-of-domain or unanswerable questions (e.g., "معنی پرچم ژاپن چیست؟" / "What is the meaning of the flag of Japan?") designed to test the system's ability to gracefully refuse to answer when the knowledge base lacks relevant information.

- **Noisy Context Questions (100 samples):** Multi-hop questions where the retrieved context is intentionally polluted with irrelevant "distractor" documents. This tests the system's robustness and its ability to identify and ignore non-pertinent information.

- **Obvious Questions (100 samples):** Simple, factoid questions for which the answer should be common knowledge within the domain (e.g., "نماز مسافر چند رکعت است؟" / "How many rak'ahs is the traveler's prayer?"). These are used to ensure the system handles simple queries correctly.

Each sample in our final evaluation dataset is structured with a question, a ground-truth answer, and category-specific metadata, providing a robust framework for the multi-faceted evaluation detailed in Section 5.

| Category | Source | Count |
|---|---|---|
| Multi-hop | IslamicPCQA | 500 |
| Negative Rejection | Manual | 100 |
| Noise Robustness | IslamicPCQA | 100 |
| Obvious | Manual | 100 |
| **Total** | | **800** |

Table 2: Composition of the Evaluation Dataset

## 4 The FARSIQA System: A FAIR-RAG Implementation

The FARSIQA system is architected as a multi-stage, agentic pipeline that implements our proposed **FAIR-RAG** framework. (Asl et al., 2025) Unlike conventional single-pass RAG systems, FARSIQA employs a dynamic, multi-step process to deconstruct, retrieve, refine, and generate answers, ensuring high levels of faithfulness and accuracy. The overall workflow, illustrated in Figure 1, can be divided into four primary phases: 1) Adaptive Query Processing, 2) Hybrid Retrieval and Re-ranking, 3) Iterative Evidence Refinement, and 4) Faithful Answer Generation. FAIR-RAG extends iterative approaches like Self-RAG (Asai et al., 2023) by incorporating checklist-based Structured Evidence Assessment (SEA).

### 4.1 Phase 1: Adaptive Query Processing

Upon receiving a user query, the system first performs an intelligent triage to determine the optimal processing path. This phase involves two key steps orchestrated by an LLM agent. (Asl et al., 2025) This task is guided by a structured prompt (see Appendix B for details).

#### 4.1.1 Query Validation and Dynamic Model Selection

The initial step is to validate the query's scope and complexity. An LLM agent classifies the input query into one of six categories. This classification serves two purposes:

**Ethical and Scope Guardrails.** Queries identified as OUT_OF_SCOPE_ISLAMIC or UNETHICAL are immediately rejected, preventing the system from engaging with irrelevant or harmful content.

**Computational Optimization.** Valid queries are categorized by complexity (VALID_OBVIOUS, VALID_SMALL, VALID_LARGE, VALID_REASONER). This allows for the dynamic allocation of computational resources. For instance, VALID_OBVIOUS questions can be answered directly by the LLM's parametric knowledge, bypassing the expensive RAG pipeline entirely. For queries requiring retrieval, an appropriately sized generator model (LLM-small or LLM-large or LLM-reasoner) is pre-selected for the final generation stage, balancing cost and performance. (Asl et al., 2025)

#### 4.1.2 Query Decomposition

For queries proceeding to the RAG pipeline, a specialized LLM agent decomposes the original question into a set of simpler, semantically distinct sub-queries (max 4). This step is critical for complex, multi-faceted, or comparative questions. For example, consider the Persian query "سهم عمده دانشمندان اسلامی در پزشکی و نجوم در دوران طلایی اسلام چه بود و تأثیر آنها بر رنسانس اروپا چه بود؟" ("What were the major contributions of Islamic scholars to medicine and astronomy during the Islamic Golden Age, and what was their influence on the European Renaissance?"). FAIR-RAG decomposes it into focused sub-queries such as "نوآوری‌های پزشکی توسط دانشمندان اسلامی مانند ابن سینا و رازی" ("Medical innovations by Islamic scholars such as Avicenna and Rhazes"), "پیشرفت‌های نجومی در رصدخانه‌های اسلامی" ("Astronomical advancements in Islamic observatories"), and "انتقال دانش علمی اسلامی به اروپا" ("The transmission of Islamic scientific knowledge to Europe"). This decomposition strategy significantly enhances retrieval recall by ensuring that all facets of the original question are targeted during the search phase. (Asl et al., 2025) This task is guided by a structured prompt (see Appendix B for details).

### 4.2 Phase 2: Hybrid Retrieval and Re-ranking

For each generated sub-query, FARSIQA employs a hybrid retrieval strategy to fetch relevant evidence from the knowledge base (described in Section 3.1).

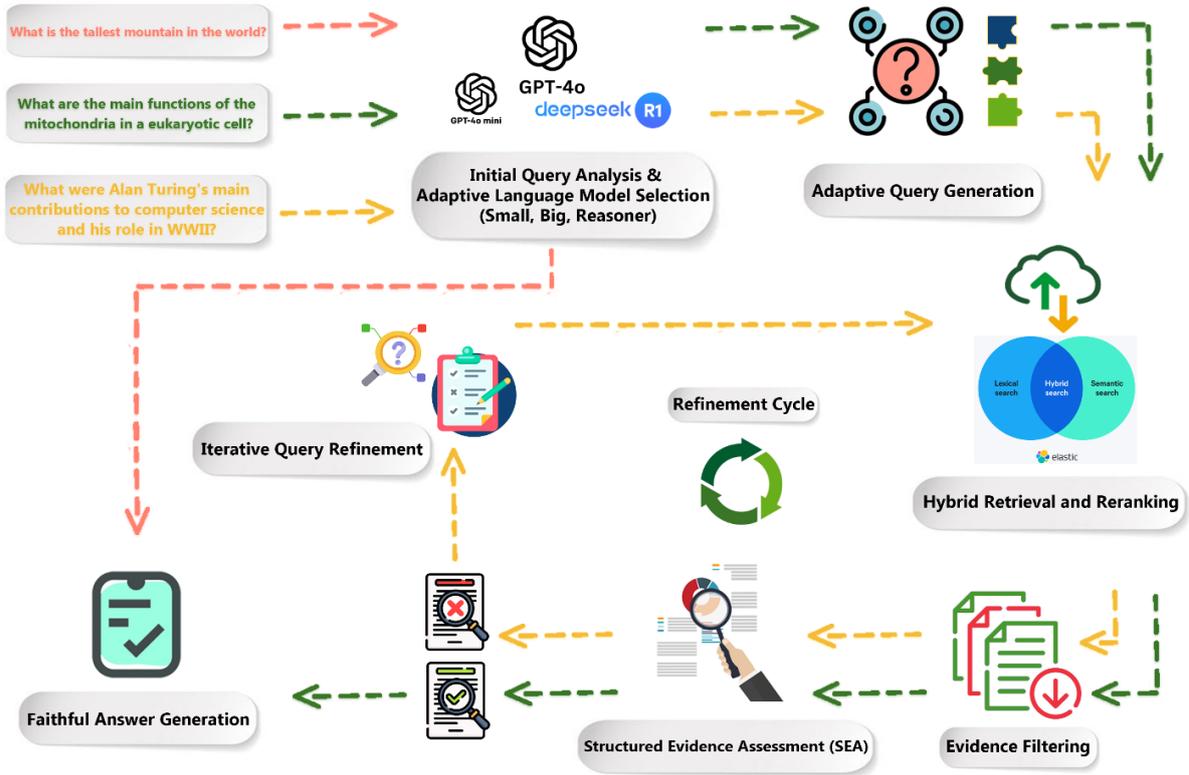

Figure 1: An overview of the multi-stage FAIR-RAG pipeline implemented in FARSIQA, showing the flow from query validation and decomposition through iterative retrieval and refinement to final answer generation.

### 4.2.1 Fine-tuned Dense Retriever

The core of our semantic search is a Transformer-based embedding model. We began with a strong pre-trained Persian model, **PartAI/Tooka-SBERT** (PartAI, 2023), and fine-tuned it on a custom dataset of **24,000 question/relevant-passage/irrelevant-passage triplets** from our Islamic corpus. Using a contrastive learning objective, the model was trained to map semantically similar queries and passages closer together in the embedding space. This domain-specific fine-tuning led to a 16% improvement in Recall@3 over the base model (detailed in Section 6). During retrieval, this model converts sub-queries into dense vectors, and Elasticsearch performs an approximate nearest neighbor search using cosine similarity to find the top-k most relevant text chunks.

### 4.2.2 Sparse Retriever

To complement the semantic search, we use a traditional sparse retrieval method (BM25) within Elasticsearch. This keyword-based search is effective at matching specific terms, names, and entities, mitigating potential failures of the dense retriever on out-of-distribution vocabulary.

### 4.2.3 Hybrid Fusion and Re-ranking

The system retrieves the top 3 documents from both the dense and sparse retrievers, resulting in a candidate set of up to 6 documents per sub-query. These candidates are then merged and re-ranked using the **Reciprocal Rank Fusion (RRF)** algorithm (Cormack et al., 2009), which effectively combines the rankings from both retrieval methods without requiring hyperparameter tuning, producing a single, robustly ranked list of the top 3 most relevant documents to pass to the next phase. We selected the top-3 documents from each retriever (dense and sparse) based on empirical tuning, leveraging the hybrid search paradigm to balance semantic depth and keyword precision. Given that user queries are of-ten decomposed into multiple sub-queries processed independently—potentially yielding diverse evidence sets—this choice optimizes the trade-off between retrieval precision, computational cost, and latency, ensuring sufficient evidence diversity without overwhelming the downstream filtering and refinement steps (as validated in ablation studies in Section 6).

## 4.3 Phase 3: Iterative Evidence Refinement (The FAIR-RAG Loop)

This phase is the core of the FAIR-RAG framework, transforming the pipeline into a deliberative reasoning process that actively seeks to build a sufficient evidence base. The loop (max 3 iterations) consists of the following steps:

**Evidence Filtering.** All retrieved documents from all sub-queries are aggregated. An LLM agent then filters this collection, identifying and discarding any documents that are irrelevant or contain only tangential information, thus reducing noise for the subsequent steps. (Asl et al., 2025) This task is guided by a structured prompt (see Appendix B for details).

**Structured Evidence Assessment (SEA).** To ensure the sufficiency and relevance of the retrieved evidence, we employ the **Structured Evidence Assessment (SEA)** methodology introduced by (Asl et al., 2025). This approach utilizes a prompted LLM agent to first deconstruct the user's query into a checklist of required informational components. It then systematically audits the retrieved documents against this checklist to identify both confirmed findings and specific "intelligence gaps." This question-centric process ensures a rigorous, targeted evaluation and provides a structured basis for deciding whether to proceed with answer generation or to initiate another retrieval cycle. This task is guided by a structured prompt (see Appendix B for details).

**Termination or Refinement.** The outcome of the Structured Evidence Assessment (SEA) dictates the pipeline's next action:

- **Termination**: If the analysis concludes that all required findings have been confirmed and there are no remaining gaps, the evidence is deemed sufficient. The iterative loop terminates, and the comprehensive evidence set is forwarded to the final answer generation phase. (Asl et al., 2025)

- **Refinement**: If the analysis identifies one or more "intelligence gaps," the system initiates a query refinement step. The detailed analysis summary, which specifies both the confirmed facts and the precise missing information, is passed to another specialized LLM agent. This agent is tasked with generating a new, highly-targeted set of sub-queries (1-4). By leveraging the already confirmed facts (e.g., using a person's identified name instead of their title), these new queries are designed to be laser-focused on resolving only the identified gaps. This strategic refinement avoids redundant searches and efficiently directs the subsequent retrieval pass to acquire the missing pieces of the puzzle. The new queries are then fed back into the Hybrid Retrieval phase (Section 4.2), and the refinement cycle continues. (Asl et al., 2025) This task is guided by a structured prompt (see Appendix B for details).

This iterative refinement process allows FARSIQA to dynamically adapt its search strategy, enabling it to solve complex, multi-hop problems that would fail in a single-pass RAG system. The maximum of three iterations was determined empirically as an optimal trade-off between response depth, latency, and computational cost.

## 4.4 Phase 4: Faithful Answer Generation

Once the refinement loop terminates with a sufficient set of evidence, the final phase generates the user-facing answer. The curated documents, along with their source URLs, are combined with the original user query into a comprehensive prompt for the generator LLM (the model selected in phase 4.1.1). The generator is governed by a strict set of instructions embedded in its prompt to ensure the final output is faithful, safe, and responsible (see Appendix B for details):

- **Strict Grounding:** The model is explicitly instructed to synthesize its answer *solely* from the provided evidence, embedding numerical citations (e.g., [1], [2]) after each piece of information to ensure full traceability.

- **Neutrality on Controversial Topics:** For sensitive or disputed theological issues, the model is guided to present differing viewpoints found in the evidence neutrally, without endorsing any single perspective.

- **Ethical Safeguards:** The system is explicitly prohibited from issuing religious edicts (*fatwas*). If a query asks for a legal ruling, the generated answer must include a disclaimer advising the user to consult a qualified religious authority. These safeguards align with ethical guidelines in AI for sensitive domains (Weidinger et al., 2021).

- **Structured Output:** The model is instructed to format its response—ranging from a short, direct answer to a comprehensive, multi-paragraph explanation with a summary—based on the depth of the available evidence.

This carefully engineered generation process is the final step in realizing the "Faithful" component of the FAIR-RAG framework, delivering a reliable and well-substantiated answer to the user. (Asl et al., 2025)

## 5 Experimental Setup

To empirically validate the effectiveness of the **FARSIQA** system and its underlying **FAIR-RAG** architecture, we designed a comprehensive evaluation framework. Our experiments are structured to assess not only the end-to-end quality of the generated answers but also the performance of individual components and the impact of our core architectural choices.

### 5.1 Evaluation Dataset

As detailed in Section 3.2, our primary evaluation is conducted on a custom benchmark of **800 questions**. This dataset is specifically curated to test the system's capabilities across a range of realistic scenarios:

- **Multi-hop Questions (500):** Multi-hop questions from the **IslamicPCQA** dataset requiring complex reasoning.

- **Negative Rejection (100):** Out-of-domain questions to test the system's ability to refuse to answer.

- **Noise Robustness (100):** Questions with intentionally noisy context to evaluate the system's filtering and focus capabilities.

- **Obvious Questions (100):** Simple, factual questions to ensure baseline correctness.

### 5.2 Evaluation Methodology: LLM-as-Judge

Given the nuanced and generative nature of our evaluation tasks, we employ an **LLM-as-Judge** methodology (Zheng et al., 2023) to ensure a scalable and consistent assessment. Our approach builds on established frameworks like G-Eval (Liu et al., 2023), utilizing meticulously designed custom prompts to mitigate known biases. For our judge, we select **Llama-4-Maverick-17B-128E-Instruct-FP8** (AI@Meta, 2025), a highly capable instruction-tuned model. It was chosen for its strong aptitude in complex instruction following and generating structured data, making it a **well-suited evaluator** for our pipeline's components. The reliability of this model for our specific, well-defined tasks is not merely an assumption; it is **empirically confirmed by a strong correlation with human expert judgments**, as we will detail in the subsequent section (5.2.1). For each evaluation aspect, a specialized prompt was designed to guide the judge model. These prompts include clear task definitions, explicit scoring criteria, and illustrative examples to ensure high-quality, structured feedback in JSON format. This methodical approach, verified by human oversight, allows for a credible and multi-faceted analysis of our pipeline (see Appendix B for details).

### 5.2.1 LLM-as-Judge is Reliable

The reliability of the Llama-4-Maverick judge for the 1-to-5 scale scoring in our ablation study was also validated. A human expert evaluated a separate set of 100 samples **drawn from the various component-level tasks**. The evaluation demonstrated a **94%** agreement between human and LLM judgments across these different tasks. This confirms the judge's capability for providing consistent, human-aligned quality assessments.

### 5.3 Evaluation Metrics

Our evaluation is organized into four key areas, assessing component-level performance, end-to-end quality, the value of iteration, and overall efficiency.

#### 5.3.1 End-to-End Quality Metrics

These metrics evaluate the final output from a user's perspective:

- **Answer Relevance:** The degree to which the generated answer directly and comprehensively addresses the user's question, rated on a scale of 1-5.

- **Answer Faithfulness (Groundedness):** Measures whether the generated answer is fully supported by the provided evidence. The judge classifies each answer as "Fully Faithful," "Partially Faithful," or "Not Faithful." We report the percentage of "Fully Faithful" answers.

- **Context Relevance:** The relevance of the final set of evidence used for generation to the original question, rated on a scale of 1-5.

- **Negative Rejection Accuracy:** The percentage of out-of-domain questions that the system correctly refused to answer.

- **Noise Robustness Accuracy:** The percentage of questions with noisy context for which the system produced a correct and robust answer, successfully ignoring the irrelevant information.

#### 5.3.2 Component-Level Performance (Ablation Insights)

To understand the internal dynamics of the FAIR-RAG pipeline, we evaluate the performance of its key modules:

- **Query Decomposition Quality:** The effectiveness of the initial query breakdown, rated on a 1-5 scale for relevance, coverage, and efficiency.

- **Document Filtering Efficacy:** Assessed via Filter Precision (proportion of kept documents that were truly relevant) and Filter Recall (proportion of relevant documents correctly kept), with relevance determined by the LLM-as-Judge methodology (Section 5.2). We also compute **F1-score** as the harmonic mean to balance these metrics.

- **Structured Evidence Assessment (SEA) Accuracy:** The accuracy of the system's analysis and decision to stop or continue the iterative loop, compared against the judge's verdict.

- **Query Refinement Quality:** The effectiveness of newly generated queries in targeting information gaps, rated on a 1-5 scale.

#### 5.3.3 Iterative Improvement

To directly measure the contribution of the iterative process, we analyze:

- **Iterative Answer Improvement:** The judge ranks the answers produced after 1, 2, 3 and 4 iterations. We report the **Improvement Rate** (the percentage of cases where the 3-iteration answer was ranked better than the 1-iteration answer) and the average rank for each iteration level.

#### 5.3.4 Efficiency Metrics

We measure the computational cost of the system via:

- **API Calls:** Average number of LLM API calls per query.

- **Token Cost:** Average number of total tokens (prompt + completion).

### 5.4 Baselines and Ablation Studies

To isolate the benefits of the FAIR-RAG architecture, we compare the full FARSIQA system against several ablated versions:

- **FARSIQA (Full System - Optimal Setting):** The complete implementation with a maximum of 3 iterations (max_iter=3), which was selected as the optimal configuration.

- **FARSIQA (Single Iteration):** The system is limited to a single retrieval and generation pass (max_iter=1), representing an advanced but non-iterative RAG pipeline.

- **FARSIQA (2 Iterations):** The system configured with a maximum of 2 iterations (max_iter=2).

- **FARSIQA (4 Iterations):** The system configured with a maximum of 4 iterations (max_iter=4).

- **Naive RAG:** A simple baseline that performs a single retrieval pass on the raw query and generates an answer without any filtering or refinement steps.

By comparing the full system against these variants, we can quantify the impact of the iterative refinement loop and the query decomposition module.

### 5.5 Implementation Details

**LLM Agents.** We use a dynamic selection of models for different tasks to optimize for cost and performance. **Llama-3-8B-Instruct** (AI@Meta, 2024) is used for less complex tasks like query decomposition and Structured Evidence Assessment (SEA). The more powerful **Llama-3.1-70B-Instruct** (AI@Meta, 2024) is used for critical tasks requiring deeper understanding, such as evidence filtering, query refinement, and final answer generation. For highly complex reasoning tasks, the system can leverage **DeepSeek-R1** (DeepSeek-AI, 2025) as a specialized reasoner model.

**Embedding Model.** Our dense retriever uses a **PartAI/Tooka-SBERT** model, fine-tuned on our custom dataset of **24k** Islamic question-passage pairs.

**Retriever Backend.** All retrieval and indexing operations are performed using **Elasticsearch 8.x**, configured for hybrid search.

This comprehensive experimental design allows for a thorough and transparent evaluation of FARSIQA's performance, providing deep insights into the effectiveness of the FAIR-RAG framework.

## 6 Results and Analysis

This section presents a comprehensive empirical evaluation of the FARSIQA system. We first report the performance of our fine-tuned retriever, then analyze its performance across multiple dimensions: end-to-end quality, component-level efficacy, the impact of iterative refinement, and computational efficiency. Our analysis includes rigorous ablation studies to demonstrate the value of each architectural component, particularly the iterative nature of the FAIR-RAG framework and our dynamic LLM selection strategy.

### 6.1 Retriever Performance

A high-performing retriever is crucial for any RAG system. Table 3 presents the performance of our dense retriever before and after fine-tuning on our custom Islamic Q&A dataset, evaluated on the IslamicPCQA benchmark.

The fine-tuned model demonstrates a clear and consistent improvement across all standard information retrieval metrics. Most notably, we observe a **13.2% increase in Mean Reciprocal Rank (MRR)** and a **16.2% increase in Recall@3**. This confirms our hypothesis that domain-specific fine-tuning effectively adapts the embedding space to the nuances of Islamic texts, enabling the model to rank relevant documents more highly. The significant uplift in top-k recall is particularly beneficial for the downstream RAG pipeline, as it provides the generator with a more accurate and concise context. While the relative improvements are significant, the absolute scores underscore the inherent difficulty of the multi-hop retrieval task in IslamicPCQA, where two distinct "golden paragraphs" must often be retrieved. This challenge, however, is precisely the scenario our iterative framework is designed to overcome.

| Model | MRR | Recall@3 | Recall@5 | Recall@10 | Prec@3 | Prec@5 | Prec@10 |
|---|---|---|---|---|---|---|---|
| Baseline | 0.2006 | 0.1110 | 0.1470 | 0.2010 | 0.0740 | 0.0588 | 0.0402 |
| Fine-tuned | **0.2271** | **0.1290** | **0.1580** | **0.2130** | **0.0860** | **0.0632** | **0.0426** |

Table 3: Performance of the Dense Retriever before (Baseline) and after Fine-Tuning.

### 6.1.1 Qualitative Analysis: Iterative Refinement as a Compensation Mechanism

To illustrate how FAIR-RAG compensates for the inherent limitations of single-pass retrieval, consider the multi-hop query: "Where was the author of the book Al-Iqtisad ila Tariq al-Rashad born?" The correct answer is "Tus". Answering this requires finding two pieces of evidence: (1) a document linking the book to its author, "Sheikh Tusi", and (2) a biographical document stating Sheikh Tusi's birthplace.

In its first iteration, FAIR-RAG generates sub-queries like "author of Al-Iqtisad…". This successfully retrieves the first golden paragraph, identifying Sheikh Tusi as the author. However, this initial, broad search fails to locate the second crucial paragraph containing his biographical details. A standard RAG system would likely fail here, lacking the necessary evidence.

This is where FAIR-RAG's iterative nature becomes critical. The Structured Evidence Assessment (SEA) module recognizes the missing information (birthplace). Armed with the newly identified entity, "Sheikh Tusi", the framework initiates a second iteration. It now generates highly targeted sub-queries such as "birth city of Sheikh Tusi" and "birthplace of Abu Ja'far Muhammad ibn Hasan Tusi". These precise queries successfully retrieve the second golden paragraph, which explicitly states he was born in Tus.

This case study serves as a compelling illustration of our core thesis: while a powerful retriever is beneficial, its inevitable failures in complex scenarios do not have to result in system-level failure. The iterative refinement cycle of FAIR-RAG acts as a critical compensation mechanism, intelligently adapting its search strategy to overcome initial retrieval weaknesses and progressively build the complete evidence base required for a faithful answer.

### 6.2 End-to-End System Performance

To quantify the overall impact of the FAIR-RAG architecture, we conducted a head-to-head comparison between the full **FARSIQA** system and a **Naive RAG** baseline. The Naive RAG baseline uses a simple retrieve-and-generate pipeline with a single LLM call and no advanced components like query decomposition, filtering, or iteration. The results, averaged across our 800-sample evaluation set, are presented in Table 4.

The results unequivocally demonstrate the superiority of the FARSIQA system. The architectural sophistication of FAIR-RAG yields substantial improvements across every evaluation dimension. We observe a significant **19-point increase in Answer Correctness Accuracy**, indicating that users are far more likely to receive a high-quality, correct answer from FARSIQA.

This is further explained by the gains in relevance metrics. **Answer Relevance** improves from 3.56 to 3.98 **(11.8% improvement)**, and **Context Relevance** from 3.31 to 3.49 **(5.4% improvement)**. These gains are directly attributable to the query decomposition and evidence filtering modules, which work in concert to first broaden the search for all facets of a query and then prune irrelevant documents, providing a cleaner, more focused context to the generator. While the improvement in **Faithfulness** is modest (1.2% improvement), it is achieved on top of a much more relevant and complex evidence base.

The most dramatic improvements are seen in the system's robustness. FARSIQA achieves a near-perfect **97.0% accuracy on Negative Rejection**, a 40-point increase over the baseline. This highlights the critical role of the initial query validation and scoping module in identifying and properly handling out-of-domain questions. Furthermore, the **16-point improvement in Noise Robustness** underscores the value of the evidence filtering and iterative refinement loops in distinguishing signal from noise, a task where the Naive RAG

| System   | Ans. Relevance (1-5) | Ans. Correct. (1-5, ≥4.0) | Faithfulness (%) | Ctx. Relevance (1-5) | Neg. Reject. (%) |
|----------|----------------------|---------------------------|------------------|----------------------|------------------|
| Naive RAG | 3.56                | 55.3%                     | 80.4%            | 3.31                 | 57.0%            |
| FARSIQA  | **3.98**             | **74.3%**                 | **81.6%**        | **3.49**             | **97.0%**        |

Table 4: End-to-End Performance of the Full FARSIQA System vs. Naive RAG Baseline. *Answer Correctness* is scored on a 1-5 scale; the reported percentage represents answers scoring ≥4.0.

system struggles.

### 6.3 Ablation Study 1: The Impact of Iterative Refinement

A core hypothesis of our work is that an iterative refinement process is essential for handling complex queries in knowledge-intensive domains. To empirically validate this and determine the optimal number of refinement cycles, we conducted an ablation study by varying the maximum number of iterations (max_iter) from 1 (a single-pass system) to 4. We evaluated the impact on three key dimensions: **end-to-end answer quality, computational cost (API calls and tokens), and response time (latency)**.

The comprehensive results are presented in Table 5. The data reveals a clear and compelling narrative: while iteration significantly boosts performance, there is a clear point of diminishing returns where additional cycles increase cost and delay without adding meaningful value.

Moving from one to three iterations yields a dramatic improvement **from 3.32 to 2.10 in the "average answer rank"** directly confirming that the iterative process is critical for gathering sufficient evidence to mitigate hallucination. The most telling statistic is the **80.1% improvement rate**, which indicates that for four out of five complex questions, the 3-iteration process produces a definitively superior answer compared to a single-pass approach.

**Analysis of Diminishing Returns (Iteration 4):** The fourth iteration marks a clear point of diminishing returns. The key quality metric—average answer rank—shows a negligible improvement from 2.10 to 2.08, a statistically insignificant gain. Intriguingly, the Improvement Rate slightly drops to 77.3%, suggesting that the fourth iteration may occasionally introduce noise that makes the final answer slightly less preferred than the more concise 3-iteration result.

This stagnation in quality is coupled with a tangible increase in both cost and latency. The fourth iteration adds nearly 7% to the average API calls and consumes approximately 900 additional tokens per query. This is accompanied by a 7.4% increase in average latency, pushing the response time from 22.1s to 23.8s. This trade-off—a notable increase in computational cost and user-perceived delay for no discernible improvement in answer quality—makes a fourth iteration clearly inefficient.

**Conclusion on Optimal Configuration:** Based on this comprehensive analysis, we establish a maximum of 3 iterations as the optimal configuration for FARSIQA. This setting maximizes answer faithfulness and relevance while maintaining an acceptable balance of **computational efficiency and response time**, delivering the full power of the FAIR-RAG framework without incurring the cost and delay of unproductive refinement cycles.

### 6.4 Ablation Study 2: The Role of Dynamic LLM Selection

A key architectural feature of FARSIQA is its dynamic allocation of different-sized LLMs for various sub-tasks. This strategy aims to balance analytical power with **financial expenditure, computational cost, and response time** by using powerful models only for the most complex reasoning steps. To validate this design, we conducted an ablation study comparing our **Dynamic** system against three static configurations: one using only a small LLM (**Static Small**), one using a large LLM (**Static Large**), and one using a specialized reasoner model (**Static Reasoner**) for all tasks.

The results, presented in Table 6, reveal a compelling trade-off between quality and efficiency, ultimately highlighting the superiority of our dynamic approach.

**Performance Analysis:**
The Static Small configuration proves inad-equate, suffering a significant **27-point drop in faithfulness** compared to our dynamic system and confirming that smaller models lack the necessary reasoning capacity for critical steps like evidence filtering. The Static Reasoner configuration achieves the highest Answer Correctness at 4.33, yet its faithfulness score is surprisingly lower than both the large and dynamic models, and it performs poorly on Negative Rejection. This suggests that while specialized for reasoning, it may be less optimized for other crucial sub-tasks like query validation or strict evidence adherence.

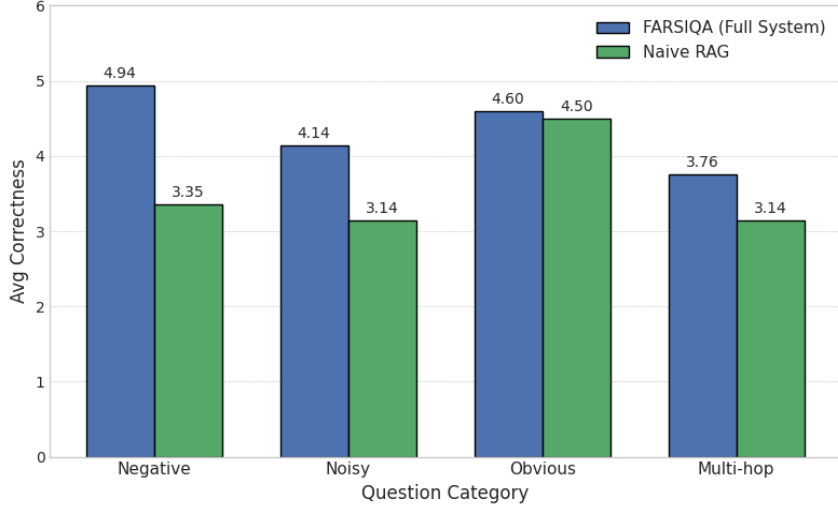

Figure 2: Answer Correctness (1-5) by Category for FARSIQA vs. Naive RAG. FARSIQA demonstrates marked superiority on challenging tasks, particularly in handling noisy contexts and out-of-scope queries.

| Max Iter. | Ans. Rank | Improve. Rate (vs. Iter 1) | API Calls | Tokens | Latency (s) |
|---|---|---|---|---|---|
| 1 (Single Pass) | 3.32 | – | 4.46 | 8,199 | 15.30 |
| 2 | 2.50 | 74.88% | 5.56 | 10,634 | 19.84 |
| 3 (Full System) | **2.10** | 80.1% | **6.07** | **11,863** | **22.137** |
| 4 | 2.08 | 77.3% | 6.48 | 12,736 | 23.77 |

Table 5: Impact of the Number of Iterations on Answer Quality and Efficiency. Answer Rank is rated by an LLM-as-Judge (lower is better). Improvement Rate is the percentage of cases where the answer was judged superior to the Iteration 1 answer.

The Static Large configuration achieves the highest faithfulness at 65.6%. However, it performs worse on **Negative Rejection (94.0% vs. 97.0%)**. This indicates that our dynamic approach—which uses a large model specifically for the crucial initial query validation step—is more effective for that task than a system using one model generically. Our dynamic system achieves a top-tier Answer Correctness score (4.06) and the best Negative Rejection score, demonstrating a more robust and well-rounded performance profile.

**Efficiency and Cost-Benefit Analysis:**

The efficiency data provides the most decisive justification for our dynamic strategy, establishing a clear superiority in cost-performance. An analysis across all configurations reveals the following:

- **Static Reasoner: Prohibitively Expensive and Impractical.** This configuration is financially unviable for any practical deployment. Its per-query cost is **over an order of magnitude (11.8x) higher** than our dynamic system. This extreme expense is coupled with an unusable average latency of 77.9 seconds, which is **3.5 times slower** than our approach. The high answer correctness score cannot justify these prohibitive operational costs and delays.

- **Static Small: A False Economy.** While being the cheapest option in absolute terms, this system represents a classic false economy. Its low cost is rendered irrelevant by its extremely **poor performance across all quality metrics**. Furthermore, it is surprisingly inefficient in terms of time, with a **38% higher latency** than our dynamic system, likely due to repeated reasoning failures and the need for additional internal processing to compensate for its limited capabilities.

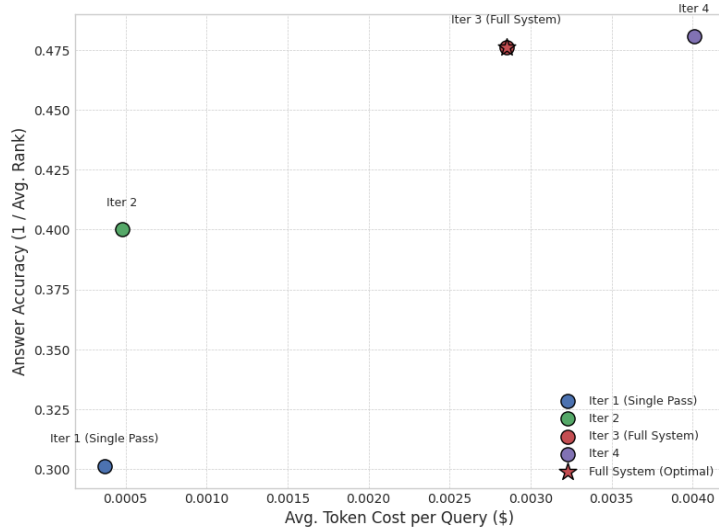

Figure 3: Trade-off between Answer Accuracy and Token Cost Across Iterations. The plot vividly depicts the principle of diminishing returns, with Iteration 3 achieving peak performance at optimal cost.

| System | Correct. (1-5) | Faithful (%) | Neg. Reject. (%) | API Calls | Tokens | Cost ($) | Latency (s) |
|---|---|---|---|---|---|---|---|
| Static Small | 3.38 | 35.4% | 74.0% | 7.94 | 16,145 | 5.33e-4 | 30.13 |
| Static Large | 4.03 | 65.6% | 94.0% | 6.07 | 11,681 | 2.89e-3 | 21.80 |
| Static Reasoner | 4.33 | 57.71% | 82% | 7.54 | 33,934 | 2.96e-2 | 77.94 |
| FARSIQA (Dynamic) | **4.06** | **62.5%** | **97.0%** | **6.07** | **11,863** | **2.51e-3** | **22.14** |

Table 6: Ablation study on LLM size. The full FARSIQA system uses dynamic allocation. All systems are run with max_iter=3.

- **Static Large vs. FARSIQA (Dynamic): The Optimal Trade-off.** The most critical comparison is against the Static Large system, where FARSIQA's intelligent design becomes evident.

  – **Cost-Effectiveness:** Our dynamic system achieves its superior and more robust performance profile (including its best-in-class 97.0% Negative Rejection rate) while being **approximately 13% more cost-effective** per query (2.51e-3 vs. 2.89e-3). This translates to significant savings at scale.
  
  – **Token and API Efficiency:** Both systems exhibit nearly identical efficiency in API calls (6.07) and token consumption (Dynamic: 11,863 vs. Large: 11,681), confirming that dynamic routing does not introduce computational overhead.
  
  – **Latency Trade-off:** The Static Large configuration is marginally faster by only 0.34 seconds. This statistically negligible 1.5% difference in latency is a minimal price to pay for the additional safeguards delivered by the dynamic system.

This establishes a clear value proposition: our dynamic system exchanges a negligible increase in latency for a **significant gain in functional robustness *and* a tangible 13% reduction in operational cost**. This makes it the unequivocally superior choice for a scalable, effective, and financially sustainable system.

**Conclusion on Optimal Configuration:** This analysis confirms the success of our dynamic LLM allocation strategy. It achieves a state-of-the-art performance profile that is statistically comparable to, and in several as-pects superior to, systems that rely exclusively on large or reasoner models. By balancing performance, financial cost, and latency, the dynamic configuration delivers near-optimal quality without the financial overhead of the Static Large system or the prohibitive expense and latency associated with the specialist model.

### 6.5 Component-Level Performance Analysis

To gain deeper insights into the internal mechanics of the FAIR-RAG pipeline, we conducted a fine-grained evaluation of its core reasoning modules. By assessing each component individually, we can identify its specific contribution to the system's overall performance and pinpoint areas for future optimization. The results of this analysis are summarized in Table 7.

**Query Decomposition:**

The initial query decomposition module achieved a high average score of **4.13 out of 5.0**. This indicates that the system is highly effective at deconstructing complex, multi-faceted user questions into a set of coherent and searchable sub-queries. This strong initial step is fundamental to the pipeline's ability to retrieve a comprehensive set of initial evidence.

**Evidence Filtering:**

The evidence filtering module demonstrates a solid balance between precision and recall, achieving an **F1-Score of 74.2%**. With a precision of 71.7%, the filter is generally successful at removing irrelevant documents from the context. However, the analysis also reveals that in its effort to reduce noise, the module can occasionally be overzealous and discard potentially useful information. This highlights a classic precision-recall trade-off that represents a key area for future tuning.

**Structured Evidence Assessment (SEA):**

The Structured Evidence Assessment (SEA) is the crucial decision-making component of the iterative loop. With an **accuracy of 66.0%** and an **F1-Score of 67.9%**, the module performs reliably better than random chance. Its **precision of 74.0%** indicates that when it decides the evidence is sufficient, it is usually correct. However, its recall of 62.6% suggests it is more prone to prematurely stopping the loop rather than continuing to search for more evidence. Improving the recall of this module could further enhance the quality of answers for the most complex queries.

**Query Refinement:**

When the system determines that the current evidence is insufficient and triggers a refinement step, its performance is exceptionally strong. The query refinement module achieved an outstanding average score of **4.61 out of 5.0**. This high score confirms that the system excels at identifying specific knowledge gaps in the existing context and generating new, highly-targeted queries to fill them. This ability to intelligently adapt its search strategy is a core strength of the FAIR-RAG architecture and a key driver of its performance on multi-hop questions.

### 6.6 A Complex Case Study: Comparative Multi-Hop Reasoning

To demonstrate the unique advantages of the FARSIQA, we analyze a hybrid comparative, multi-hop query from the Islamic domain. This type of query is particularly challenging because it requires the system to conduct two parallel lines of multi-hop reasoning and then synthesize the results into a coherent comparison.

**The query is:** "محل دفن پیامبری که توسط نهنگ بلعیده شد را با شهری که پیامبری که کعبه را ساخت در آن متولد شد، مقایسه کنید." ("Compare the burial place of the Prophet who was swallowed by a whale with the city where the Prophet who built the Kaaba was born.")

**Why This Query is Difficult for Standard RAG Frameworks:**

A standard RAG system would likely fail because it treats the query as a single, overloaded semantic vector. It would struggle to simultaneously resolve two separate, multi-step entities ("Prophet swallowed by a whale" → Yunus → Nineveh) and ("Prophet who built the Kaaba" → Ibrahim → Ur). The system would likely retrieve a document about one Prophet but fail to find the other, or retrieve general documents that lack the specific geographical details required.

**FAIR-RAG in Action:**

| Component | Metric | Value |
|---|---|---|
| Query Decomposition | Avg. Score (1-5) | 4.13 |
| Evidence Filtering | Precision | 71.7% |
| | Recall | 76.8% |
| | F1-Score | 74.2% |
| Structured Evidence Assessment (SEA) | Accuracy | 66.0% |
| | Precision | 74.0% |
| | Recall | 62.6% |
| | F1-Score | 67.9% |
| Query Refinement | Avg. Score (1-5) | 4.61 |

Table 7: Performance of FARSIQA's Core Internal Modules. Scores are evaluated by an LLM-as-Judge.

- **Iteration 1: Semantic Decomposition & Parallel Initial Retrieval**
  - **Adaptive Sub-Queries:** FAIR-RAG's first action is to decompose the comparative query into two distinct, parallel investigative tracks:
    * Track A: ["محل دفن پیامبری که توسط نهنگ بلعیده شد"] ("burial place of the Prophet swallowed by a whale")
    * Track B: ["زادگاه پیامبری که کعبه را ساخت"] ("birth city of the Prophet who built the Kaaba")
  - **Retrieved Evidence:** The system retrieves initial evidence for both tracks concurrently:
    * Evidence 1: "حضرت یونس پیامبری است که به عنوان آزمایشی از سوی خدا توسط یک ماهی بزرگ (یا نهنگ) بلعیده شد و پس از نجات به میان قوم خود بازگشت تا دعوت الهی را ادامه دهد." (*"Prophet Yunus (Jonah) is the prophet who was famously swallowed by a large fish (or whale) as a trial from God. After being saved, he returned to his people to preach."*)
    * Evidence 2: "حضرت ابراهیم (ع) به همراه پسرش اسماعیل پایه‌های کعبه را در مکه بنا نهاد تا خانه‌ای برای عبادت خدا باشد." (*"Prophet Ibrahim (Abraham), with the help of his son Ismail, is credited with constructing the foundations of the Kaaba in Mecca as a house of worship for God."*)
  - **Structured Evidence Assessment (SEA):**
    * **is_sufficient:** 'No'
    * **analysis_summary:** The initial analysis successfully identified the primary entities for both comparative tracks: Prophet Yunus and Prophet Ibrahim. However, the key required findings regarding the associated geographical locations (burial place and birthplace) have not yet been addressed. Significant information gaps remain, precluding a complete answer.

- **Iteration 2: Parallel Query Refinement & Evidence Completion**
  - **Refined Queries:** The refinement module now uses the entities identified in Iteration 1 to generate new, highly-focused queries for each track:
    * Refined Query A: ["آرامگاه حضرت یونس"] ("tomb of Prophet Yunus")
    * Refined Query B: ["زادگاه حضرت ابراهیم"] ("birthplace of Prophet Ibrahim")
  - **Retrieved Evidence:** The new targeted queries retrieve the final missing pieces of information:
    * Evidence 3: "محل سنتی آرامگاه حضرت یونس بر روی تپه‌ای در شهر باستانی نینوا، نزدیک موصل امروزی در عراق، قرار دارد." (*"The traditional site of the tomb of Prophet Yunus is located on a hill in the ancient city of Nineveh, which is near modern-day Mosul, Iraq."*)
    * Evidence 4: "منابع تاریخی و مذهبی نشان می‌دهند که حضرت ابراهیم در شهر باستانی اور کلدانیان در جنوب عراق کنونی متولد شده است." (*"Historical and religious sources indicate that Prophet Ibrahim was born in the

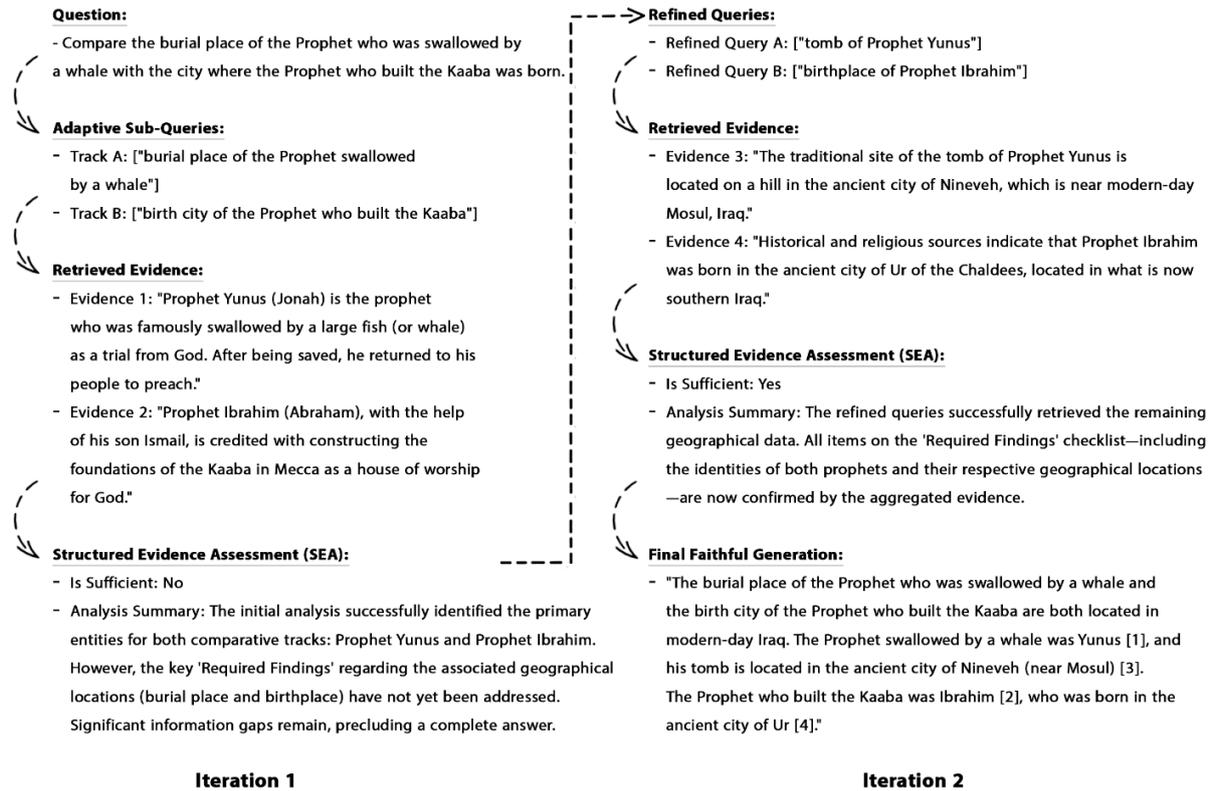

Figure 4: Qualitative case study of iterative refinement in FAIR-RAG. The system systematically decomposes a complex comparative query, pursues parallel reasoning tracks, and applies targeted refinement to fill information gaps before synthesizing a comprehensive answer.

   *ancient city of Ur of the Chaldees, located in what is now southern Iraq."*)

- **Structured Evidence Assessment (SEA):**
  * **is_sufficient:** 'Yes'
  * **analysis_summary:** The refined queries successfully retrieved the remaining geographical data. All items on the 'Required Findings' checklist—including the identities of both prophets and their respective geographical locations—are now confirmed by the aggregated evidence. No information gaps remain, enabling a direct, evidence-grounded comparison.

- **Final Faithful Generation:** The system synthesizes the evidence from both parallel tracks into a single, structured, and fully-grounded comparative answer:

"محل دفن پیامبری که توسط نهنگ بلعیده شد و شهر زادگاه پیامبری که کعبه را ساخت، هر دو در عراق در شهر باستانی [1] حضرت یونس امروزی قرار دارند. به خاک سپرده شده است، [3] نینوا (نزدیک موصل) به دنیا [4] اور در شهر باستانی [2] حضرت ابراهیم و آمد." ("The burial place of the Prophet who was swallowed by a whale and the birth city of the Prophet who built the Kaaba are both located in modern-day Iraq. The Prophet swallowed by a whale was **Yunus** [1], and his tomb is located in the ancient city of **Nineveh (near Mosul)** [3], while the Prophet who built the Kaaba was **Ibrahim** [2], who was born in the ancient city of **Ur** [4].")

This case study highlights FAIR-RAG's key architectural advantage: its ability to systematically decompose a complex goal into parallel sub-problems and then apply iterative refinement to solve each one before synthesizing a comprehensive final answer. This structured, multi-threaded reasoning process is what en-ables it to succeed where other advanced RAG frameworks may fail.

### 6.7 Failure Mode Analysis

To move beyond aggregate performance metrics and gain a deeper, qualitative understanding of FARSIQA's limitations, we conducted a rigorous failure mode analysis. This process adopted a hybrid human-AI methodology, using an advanced LLM judge for initial categorization, followed by meticulous expert human validation. We analyzed a comprehensive set of 122 unique underperforming queries from our test set. For each failure case, we supplied the complete execution trace (including intermediate agent outputs) to the judge, which was guided by a detailed taxonomy and structured prompt (Appendix G). Human reviewers then audited every prediction to ensure contextual fidelity before final labeling.

The overall results of this analysis are visualized in Figure 5. The distribution of primary error modes reveals that while failures occur across the pipeline, they are heavily concentrated in the final two stages. Generation Failures emerged as the most dominant error category, accounting for a significant majority of all cases (54.9%). Retrieval Failures were the second most common at 27.9%. Errors in the upstream control flow were comparatively rare: Query Decomposition Errors accounted for 9.0%, Evidence Filtering Errors for 5.7%, and Structured Evidence Assessment (SEA) Errors for 2.5%. Notably, we observed no instances of Query Refinement Error, indicating that once an information gap is surfaced, the refinement module generally succeeds in targeting it.

**Generation Failures (54.9%).** Representing over half of all errors, this category is unequivocally the primary bottleneck in the FARSIQA architecture. These failures occur when the system successfully retrieves and filters the correct evidentiary documents but fails to synthesize them into a factually accurate answer. We identified several recurring root causes:

- **Flawed Logical Inference:** The model struggles with reasoning tasks that require understanding implicit relationships, particularly with cyclical concepts or complex relational chains.

- **Misinterpretation of Question Intent:** The generator often fails to adhere strictly to the user's specific query, instead providing a related but incorrect answer.

- **Incorrect Entity Relationship Mapping:** A frequent sub-type of flawed inference where the model fails to correctly map complex relationships described in the query.

- **Ignoring Correct Evidence:** In cases with conflicting or noisy evidence, the model sometimes grounds its answer on an incorrect document while ignoring the document containing the correct information.

**Retrieval Failures (27.9%).** As the second-largest category, these errors represent a fundamental inability to source the required information from the knowledge base. The predominant cause was Knowledge Base Gaps, where information for highly specific, long-tail queries was entirely absent from the corpus. A secondary contributor was the occasional generation of ineffective sub-queries that were too general to isolate the required fact, especially for nuanced historical details.

**Query Decomposition Errors (9.0%).** These initial-stage failures arose when the system misunderstood the semantic intent of the user's request. The primary root cause was an inability to correctly parse compound relational constraints (for example, ordinal relationships such as "the third son of..."), leading to overly broad or misdirected sub-queries that derailed downstream retrieval.

**Evidence Filtering & SEA Errors (8.2% combined).** Intermediate-stage errors were relatively infrequent, suggesting that the pipeline's control flow is generally robust. Filtering errors largely stemmed from over-aggressive pruning that discarded a relevant document. SEA errors reflected premature sufficiency judgments, where the loop terminated before all checklist items had supporting evidence.

In summary, this comprehensive failure analysis reveals that FARSIQA's primary vulnerability lies not in its ability to find information, but in its capacity to reason over and faithfully synthesize it. The overwhelming prevalence of Generation Failures highlights the final language model as the most critical area for future research. While improving knowledge base coverage can mitigate retrieval issues, the most substantial gains are likely to come from enhancing the generator's evidence adherence and complex reasoning capabilities.

## 7 Conclusion and Future Work

In this paper, we addressed the critical challenge of developing a reliable and faithful question-answering system for the high-stakes, nuanced domain of Persian Islamic studies. We introduced **FARSIQA**, a novel QA system built upon our **FAIR-RAG** architecture—a framework designed for Faithful, Adaptive, Iterative Refinement. Our work represents a significant step forward in creating specialized AI systems that can handle complex queries in sensitive domains with a strong emphasis on accuracy and traceability.

Our primary contribution is twofold: to the best of our knowledge, the first robust, end-to-end system in Persian Islamic QA, and the introduction of an advanced RAG architecture optimized for complex reasoning. Through extensive experiments (as detailed in Section 6), we demonstrated that the iterative refinement loop at the core of FAIR-RAG is crucial for improving answer quality. Our full system achieved a **negative rejection accuracy of 97.0%** (a 40-point improvement over baselines), validating that the query validation and iterative evidence gathering processes are vital for robustly handling out-of-scope or harmful queries in sensitive domains. Furthermore, by creating a comprehensive knowledge base of over one million documents and releasing a fine-tuned, domain-specific retriever model, we have established a strong new benchmark for future research in Persian NLP and digital humanities.

### 7.1 Limitations and Future Work

Despite its strong performance, FARSIQA has several limitations that open avenues for future research.

- **Lack of Conversational Context:** The current system is stateless and processes each query independently. It lacks a conversational memory, preventing it from understanding follow-up questions or retaining context from a user's dialogue history.

- **Knowledge Base Coverage:** While extensive, our knowledge base does not encompass the entirety of Islamic scholarly texts. Expanding it to include a wider range of sources, particularly classical hadith collections and diverse jurisprudential opinions, would further enhance its comprehensiveness. Additionally, potential cultural or interpretive biases in the curated sources could affect representation of minority views.

- **Latency:** The multi-step, iterative nature of the FAIR-RAG pipeline, while effective, introduces latency (averaging 6.07 API calls per query, as shown in Section 6.3), which may constrain real-time applications.

- **Acknowledgment of Bias:** We have endeavored to incorporate a diverse range of Islamic perspectives to build a balanced knowledge base. However, for the sake of full transparency, it is crucial to acknowledge the composition of our current knowledge corpus, a comprehensive list of which is provided in Table 1. An analysis of these sources reveals that the collection predominantly features texts from the Shi'a school of Islamic thought. Consequently, while the FARSIQA system is designed to generate faithful and neutral responses based on the provided context, the answers may inherently reflect the theological and jurisprudential perspectives dominant within the source material. This represents a known limitation of the current system. We recognize the importance of incorporating a broader spectrum of Islamic schools of thought in future iterations to further mitigate this potential bias and enhance the system's universality.

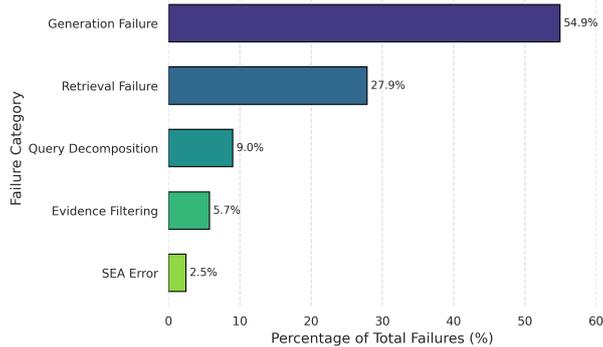

Figure 5: Distribution of primary failure modes in the FARSIQA system across 122 failed queries. Generation Failures represent the most significant bottleneck, accounting for over half of all observed errors, followed by Retrieval Failures.

To address these limitations, our future work will proceed in several directions. First, we plan to integrate a **conversational memory module** using techniques like context caching and chain-of-thought prompting (Wei et al., 2022) to enable multi-turn dialogues. Second, we will explore methods for **knowledge base expansion** and **continuous updates**, such as automated crawling and verification of new scholarly sources. Third, investigating model optimization techniques such as **quantization** (Dettmers et al., 2023) and distillation could help reduce latency without significantly compromising quality. Finally, developing a mechanism for incorporating a **user feedback loop** would allow the system to learn from its mistakes and improve over time, potentially through reinforcement learning from human feedback (RLHF).

In conclusion, this work demonstrates that by moving beyond simple retrieve-and-read pipelines, it is possible to build specialized QA systems that are not only powerful but also trustworthy. Beyond technical advancements, FARSIQA promotes equitable access to Persian Islamic knowledge, fostering cultural preservation and informed discourse. We believe that frameworks like FAIR-RAG can serve as a blueprint for developing responsible AI in other high-stakes domains, such as law and medicine, where accuracy and faithfulness are paramount.

## References

AI@Meta. 2024. The llama 3 herd of models. *arXiv preprint arXiv:2407.12345*.

AI@Meta. 2025. Llama 4: Maverick language models. Unpublished technical report.

Ahmet Yusuf Alan, Enis Karaarslan, and Ömer Aydin. 2025. A rag-based question answering system proposal for understanding islam: Mufassirqas llm.

Akari Asai, Zeqiu Wu, Yizhong Wang, Avirup Sil, and Hannaneh Hajishirzi. 2023. Self-rag: Learning to retrieve, generate, and critique through self-reflection. *arXiv preprint arXiv:2310.11511*.

Mohammad Aghajani Asl, Majid Asgari-Bidhendi, and Behrooz Minaei-Bidgoli. 2025. Fair-rag: Faithful adaptive iterative refinement for retrieval-augmented generation.

Tom B. Brown, Benjamin Mann, Nick Ryder, Melanie Subbiah, Jared Kaplan, Prafulla Dhariwal, Arvind Neelakantan, Pranav Shyam, Girish Sastry, Amanda Askell, Sandhini Agarwal, Ariel Herbert-Voss, Gretchen Krueger, Tom Henighan, Rewon Child, Aditya Ramesh, Daniel M. Ziegler, Jeffrey Wu, Clemens Winter, Christopher Hesse, Mark Chen, Eric Sigler, Mateusz Litwin, Scott Gray, Benjamin Chess, Jack Clark, Christopher Berner, Sam McCandlish, Alec Radford, Ilya Sutskever, and Dario Amodei. 2020. Language models are few-shot learners. *Advances in Neural Information Processing Systems*, 33:1877–1901.

Danqi Chen, Adam Fisch, Jason Weston, and Antoine Bordes. 2017. Reading wikipedia

to answer open-domain questions. *arXiv preprint arXiv:1704.00051*.

Gordon V Cormack, Charles LA Clarke, and Stefan Buettcher. 2009. Reciprocal rank fusion outperforms condorcet and individual rank learning methods. In *Proceedings of the 32nd international ACM SIGIR conference on Research and development in information retrieval*, pages 758–759.

DeepInfra, Inc. 2025. DeepInfra API Pricing. https://deepinfra.com/pricing. Accessed: October 2025.

DeepSeek-AI. 2025. Deepseek-r1: An open large reasoning model family. *arXiv preprint arXiv:2501.05624*.

Tim Dettmers, Artidoro Pagnoni, Ari Holtzman, and Luke Zettlemoyer. 2023. Qlora: Efficient finetuning of quantized llms. *arXiv preprint arXiv:2305.14314*.

H Fani, F Zarrinkalam, E Bagheri, and W Du. 2021. A survey on persian question answering systems. *ACM Computing Surveys*.

Mehrdad Farahani, Mohammad Gharachorloo, Marzieh Farahani, and Mohammad Manthouri. 2021. Parsbert: Transformer-based model for persian language understanding. *Neural Processing Letters*, 53(6):3831–3847.

Yunfan Gao, Yun Xiong, Xinyu Gao, Kangxiang Jia, Jinliu Pan, Yuxi Bi, Yi Dai, Jiawei Sun, and Haofen Sun. 2023. Retrieval-augmented generation for large language models: A survey. *arXiv preprint arXiv:2312.10997*.

Arash Ghafouri, Mohammad Aghajani Asl, Hasan Naderi, and Mahdi Firouzmandi. 2025. Islamicpcqa: A dataset for persian multi-hop complex question answering in islamic text resources. *IEEE Transactions on Audio, Speech and Language Processing*, 33:3801–3812.

Ziwei Ji, Nayeon Lee, Rita Frieske, Tiezheng Yu, Dan Su, Yan Xu, Etsuko Ishii, Yejin Bang, Andrea Madotto, and Pascale Fung. 2023. Survey of hallucination in natural language generation. *ACM Computing Surveys*, 55(12):1–38.

Hao Jiang, Qian Chen, Junkun He, Chin-Yew Lin, Yelong Lyu, and Xin Zhang. 2023a. Query rewriting for retrieval-augmented large language models. *arXiv preprint arXiv:2305.14283*.

Zhengbao Jiang, Vladimir Lialin, Carroll Lin, Jane Liu, and Yelong Cheng. 2023b. Flare: Forward-looking active retrieval augmented generation. *arXiv preprint arXiv:2305.06983*.

Dan Jurafsky and James H. Martin. 2023. Speech and Language Processing (3rd ed. draft).

Vladimir Karpukhin, Barlas Oguz, Sewon Min, Patrick Lewis, Ledell Wu, Sergey Edunov, Danqi Chen, and Wen-tau Yih. 2020. Dense passage retrieval for open-domain question answering. *arXiv preprint arXiv:2004.04906*.

Mojtaba Komeili, Kurt Shuster, and Jason Weston. 2022. Internet-augmented dialogue generation. *arXiv preprint arXiv:2107.07566*.

Patrick Lewis, Ethan Perez, Aleksandra Piktus, Fabio Petroni, Vladimir Karpukhin, Rodrigo Nogueira, Heinrich Paux, Pontus Stenetorp, Timo Rocktäschel, Sebastian Riedel, et al. 2020. Retrieval-augmented generation for knowledge-intensive nlp tasks. *Advances in Neural Information Processing Systems*, 33:9459–9474.

Yang Liu, Dan Iter, Yichong Xu, Shuohang Wang, Ruochen Xu, and Chenguang Zhu. 2023. G-eval: Nlg evaluation using gpt-4 with better human alignment. *arXiv preprint arXiv:2303.16634*.

S Mehraban and M Rahmati. 2022. Peransel: Answer selection model for persian question answering. *Computer Science & Engineering*.

PartAI. 2023. Tooka-sbert-v1. https://huggingface.co/PartAI/Tooka-SBERT-v1.

Ori Ram, Yoav Levine, Itay Dalmedigos, Dor Muhlgay, Amnon Shashua, Kevin Leyton-Brown, and Yoav Shoham. 2023. In-context

retrieval-augmented language models. *arXiv preprint arXiv:2302.00083*.

Stephen Robertson and Hugo Zaragoza. 2009. The probabilistic relevance framework: Bm25 and beyond. *Foundations and Trends® in Information Retrieval*, 3(4):333–389.

Jason Wei, Xuezhi Wang, Dale Schuurmans, Maarten Bosma, Fei Xia, Ed Chi, Quoc V Le, Denny Zhou, et al. 2022. Chain-of-thought prompting elicits reasoning in large language models. *Advances in Neural Information Processing Systems*, 35:24824–24837.

Laura Weidinger, John Mellor, Maribeth Rauh, Conor Griffin, Jonathan Uesato, Po-Sen Huang, Myra Cheng, Mia Glaese, Borja Balle, Atoosa Kasirzadeh, et al. 2021. Ethical and social risks of harm from language models. *arXiv preprint arXiv:2112.04359*.

Zhilin Yang, Peng Qi, Saizheng Zhang, Yoshua Bengio, William W Cohen, Ruslan Salakhutdinov, and Christopher D Manning. 2018. Hotpotqa: A dataset for diverse, explainable multi-hop question answering. *arXiv preprint arXiv:1809.09600*.

Shunyu Yao, Jeffrey Zhao, Dian Yu, Nan Du, Izhak Shafran, Karthik Narasimhan, and Yuan Cao. 2022. React: Synergizing reasoning and acting in language models. *arXiv preprint arXiv:2210.03629*.

Lianmin Zheng, Wei-Lin Chiang, Ying Sheng, Siyuan Zhuang, Zhanghao Wu, Yonghao Zhuang, Zi Lin, Zhuohan Li, Dacheng Brooks, Eric Xing, et al. 2023. Judging llm-as-a-judge with mt-bench and chatbot arena. *arXiv preprint arXiv:2306.05685*.

## A  Knowledge Base Source Details

This appendix provides a detailed description of the encyclopedic and Q&A platform sources that constitute the knowledge base used in this research, as summarized in Table 1.

### A.1  Encyclopedic Sources

This section details the encyclopedias and their respective areas of focus.

- **WikiShia:** A specialized encyclopedia focusing on topics related to Shia Islam, encompassing beliefs, personalities, literature, significant locations, historical events, rituals, and sects. URL: `https://fa.wikishia.net/`

- **WikiFeqh:** A comprehensive resource covering concepts and topics within the Islamic sciences, such as Quranic exegesis (Tafsir), jurisprudence (Fiqh), principles of jurisprudence (Usul al-Fiqh), philosophy, and theology (Kalam). URL: `https://fa.wikifeqh.ir/`

- **WikiAhlolbait (Dāneshname-ye Eslāmī):** A comprehensive encyclopedia covering a wide range of subjects in Islam. URL: `https://wiki.ahlolbait.com/`

- **ImamatPedia:** An encyclopedia dedicated to the concepts of Imamate (leadership) and Wilayat (guardianship), detailing the lives of the Infallibles (Ma'sumin), related historical events, and key figures. URL: `https://fa.imamatpedia.com/`

- **IslamPedia:** A general-purpose encyclopedia on various Islamic subjects. URL: `http://www.islampedia.ir/`

- **WikiHaj:** A specialized resource for terminology and concepts related to the Hajj (the major Islamic pilgrimage to Mecca) and Ziyarah (pilgrimage to holy sites). URL: `https://wikihaj.com/`

- **WikiNoor:** A comprehensive digital encyclopedia covering a broad spectrum of Islamic knowledge, often providing access to digitized books and articles. URL: `https://fa.wikinoor.ir/`

- **WikiPasokh:** An encyclopedia structured in a question-and-answer format, addressing topics in Quranic sciences, theology (Kalam), law, comparative religion, ethics, and history. URL: `https://fa.wikipasokh.com/`

- **WikiHussain:** A thematic encyclopedia centered on the life, teachings, and legacy of Imam Hussain ibn Ali. URL: `https://fa.wikihussain.com/`

- **The Great Islamic Encyclopedia (Dā'erat-ol-Ma'āref-e Bozorg-e Eslāmī):** A major, authoritative scholarly encyclopedia covering a vast range of topics related to Islamic civilization, history, and culture. URL: https://www.cgie.org.ir/

- **Qomnet.johd.ir:** A broad digital library and resource hub hosted by the Jihad-e Daneshgahi (Academic Center for Education, Culture and Research) of Qom, covering diverse topics in Islamic studies. URL: https://qomnet.johd.ir/

### A.2 Question-Answering (Q&A) Platforms

This section lists the Q&A platforms, which primarily contain collections of questions posed by users and authoritative answers provided by Islamic scholars and institutions.

- **IslamQuest.net:** https://www.islamquest.net/fa/
- **rasekhoon.net:** https://rasekhoon.net/
- **porseman.com:** https://porseman.com/
- **aminsearch.com:** https://aminsearch.com/
- **makarem.ir:** https://www.makarem.ir/
- **hawzah.net:** https://hawzah.net/
- **bahjat.ir:** https://bahjat.ir/
- **pasokh.org:** https://pasokh.org/
- **al-khoei.us:** https://al-khoei.us/
- **pasokhgoo.ir:** https://pasokhgoo.ir/
- **porsemanequran.com:** http://porsemanequran.com/
- **islamqa.com:** http://islamqa.com/

## B  FAIR-RAG Pipeline Prompts

This appendix presents the complete prompts that steer each agent in the FAIR-RAG pipeline. We reproduce the exact instructions used in our implementation to facilitate replication and downstream analysis. Every prompt is intentionally verbose, as the formatting (including Markdown markers and explicit examples) materially affects model behavior.

### B.1  Query Validation and Dynamic Model Selection Prompt

The initial validation agent uses the following prompt to verify scope and ethics before selecting the appropriate model size for downstream execution.

---

PROMPT = """

**Situation:** A user has submitted a question to an Islamic Question Answering System.

**Intent:**

1. Determine if the user's question is within the system's Islamic knowledge scope and adheres to general ethical guidelines.

2. If the question is valid, assess its complexity and determine the most appropriate language model (SMALL, LARGE, or REASONER) to process it, with a strong preference for the LARGE model.

**Scaffolding:**

You are provided with the user's question, rules for scope/ethics, and guidelines for model selection.

Analyze the question and classify it by outputting ONLY one of the following exact labels after "Selected Label:":

- "VALID_OBVIOUS": If the question is related to Islamic knowledge, is ethical, AND is so obvious that the model can confidently answer it without retrieval, purely from general common knowledge

---

(e.g., ``قبله مسلمانان به کدام سمت است؟''، ``ماه رمضان چند روز است؟'').

---

[Translate: (e.g., "In which direction do Muslims face during prayer (Qibla)?", "How many days does the month of Ramadan last?").]

- "VALID_SMALL": If the question is related to Islamic knowledge, is ethical, AND is very simple

---

``like recall factual direct (e.g., فرق بین نمازهای واجب (فرض) و مستحب (سنت) چیست؟``).

[Translate: (e.g., direct factual recall like "What is the difference between obligatory (Fard) and recommended (Sunnah) prayers?").]

- "VALID_LARGE": If the question is related to Islamic knowledge, is ethical, AND requires explanation, interpretation, comparison, or nuanced understanding. THIS IS THE PREFERRED DEFAULT for most valid questions

(e.g., ``توضیح مفهوم توحید در قرآن.``, ``تفاوت بین حدیث صحیح و ضعیف چیست؟``).

[Translate: (e.g., "Explain the concept of Tawhid (the oneness of God) in the Qur'an.", "What is the difference between a Sahih (authentic) and a Da'fi (weak) Hadith?").]

- "VALID_REASONER": If the question is related to Islamic knowledge, is ethical, AND specifically requires multi-step logical deduction, complex rule application, or mathematical calculations based on Islamic principles

(e.g., ``شخصی فوت کرده و یک همسر، دو پسر و یک دختر دارد. اگر ماترک او ۱۲۰ سکه طلا باشد، سهم‌الارث هر یک چقدر است؟``). Use this very sparingly.

[Translate: (e.g., "A person passed away leaving behind a spouse, two sons, and a daughter. If the estate amounts to 120 gold coins, what is the inheritance share of each heir?"). Use this very sparingly.]

- "OUT_OF_SCOPE_ISLAMIC": If the question is ethical and has **no anchor** to any specific Islamic personality, book, concept, or event (e.g., "How to configure a router?").

- "UNETHICAL": If the question conflicts with general ethical guidelines or promotes harmful content.

**Rules for Classification:**

1. **Ethical Boundaries:**
   - Do NOT process questions that conflict with general ethical guidelines or promote harmful content.

2. **Islamic Scope (Coverage of Response):**
   - **This is the most important rule: If the question is anchored to a specific Islamic personality, book, concept, historical event, or place, it MUST be considered IN SCOPE.**
   - Only reject questions that have absolutely no connection to Islamic knowledge, history, civilization, culture, or related fields.
   - **Hybrid Questions Example:** A query like

``مولف کتاب حقیقة مصحف فاطمة عند الشیعة مدرک دکتری در چه رشته ای دریافت کرد؟``

(What PhD did the author of 'The Truth of Mushaf Fatima' receive?) **MUST be classified as "VALID_LARGE"**. Even though it asks about an academic degree (a general knowledge fact), it is anchored to a specific Islamic author and book, making it valid.

3. **Model Selection Preference:**
   - **Default to "VALID_LARGE"** for any valid Islamic question unless it's extremely simple or explicitly requires complex reasoning/calculation.
   - Use "VALID_SMALL" only if the question is exceptionally straightforward and requires minimal processing.
   - Use "VALID_REASONER" only if the question cannot be adequately addressed by a standard large model and distinctly involves step-by-step calculations or complex logical chaining (like inheritance).

**User Question:** "{user_query}"

**Constraints:**

- Respond with ONLY one of the six labels listed above.
- Do not provide any explanations or additional text.
- The label MUST be on a new line after "Selected Label:".

**Output:**

Selected Label: """

## B.2 Query Decomposition Prompt

This agent decomposes complex questions into focused retrieval targets. The explicit worked example included in the prompt proved critical for consistent behavior across model checkpoints.

PROMPT = """

**Situation:** You are an expert query analyst for an Islamic knowledge Question-Answering system. A user has asked a question that might be complex, comparative, or multi-faceted. Your task is to decompose this question into a set of precise, meaningful, and distinct sub-queries to ensure the retrieval system can find comprehensive and accurate evidence from a database.

**Intent:** Decompose the original user question into its core semantic components. Transform these components into short, keyword-rich, and meaningful search phrases in Persian. The goal is to generate queries that, when searched, will collectively **cover all aspects of the original question**.

**Scaffolding:**

First, understand the principles of effective decomposition:

1. **Identify Distinct Concepts:** Separate the main subjects, actions, conditions, and comparisons in the query.
2. **Use Synonyms & Related Terms:** Think about different ways a concept might be phrased in the database (e.g., "The term 'interaction' can be searched as 'cooperation' or 'relationship'.(".

3. **Create Meaningful Phrases:** Instead of single keywords, generate short phrases that preserve the context of the sub-question.
4. **Cover All Angles:** Ensure every part of the original question is represented by at least one sub-query.

Now, study the following example carefully to understand how to apply these principles.

--- EXAMPLE ---

Original User Query: ``متفکران اسلامی در طول تاریخ چگونه مفهوم عدالت در قرآن را تفسیر کرده و آن را در حکومت به کار برده‌اند؟''

[Translate: "How have Islamic thinkers historically interpreted the concept of justice in the Quran and applied it to governance?"]

**Rationale/Analysis (This is your thought process):**

The question contains two primary components:

1. تفسیر عدالت در قرآن - تفسیر و شرح عدالت

[Translate: Interpretation of justice in the Quran - scholarly exegesis and commentary on the concept of justice ('Adālah').]

2. کاربرد آن در حکومت - رویه‌های تاریخی و اندیشه سیاسی

[Translate: Application to governance - historical implementations and reflections in Islamic political thought.]

A good search needs to find evidence for main aspects separately.

**Optimized Queries (Output):**

1. تفسیر مفهوم عدالت در قرآن توسط متفکران اسلامی

[Translate: Exegesis of the concept of justice in the Quran by Islamic scholars]

2. عدالت در حکومت از منظر فقه و فلسفه سیاسی اسلامی

[Translate: The notion of justice in governance from the perspective of Islamic jurisprudence and political philosophy]

3. تحلیل تاریخی کاربرد عدالت قرآنی در مدیریت جامعه

[Translate: Historical analysis of the application of Quranic justice in societal administration]

4. اندیشه سیاسی اسلامی و مفهوم عدالت

[Translate: Islamic political thought and the conceptualization of justice]

--- END OF EXAMPLE ---

Now, apply this exact methodology to decompose the following query.

**User Query:** "{user_query}"

**Constraints:**

- The output must be a list of meaningful search phrases.
- Each phrase must be on a new line, prefixed with a hyphen (-).
- Queries must be in Persian.
- Generate an optimized list of 1 to 4 sub-queries. Create ONLY as many as are **truly necessary** to cover all aspects of the original question.

**Output:** (just write Optimized Queries and do not explain any more and do not say "Here are the optimized queries:" or something like that.)

Optimized Queries: (A list of optimized queries)

"""

### B.3 Evidence Filtering Prompt

The filtering agent examines batched retrieval results and discards low-value passages while preserving any chunk that might carry useful facts.

PROMPT = """

**Situation:** A batch of candidate evidence documents has been retrieved for a user's query. To generate a precise and focused answer, we must filter out any documents that do not provide substantial, related information.

**Intent:** Identfiy and list the temporary IDs of all "Not Useful" documents. A document is considered "Not Useful" fi it is completely irrelevant, off-topic, or only tangentially related and **does not contain specfiic facts or information needed to answer the user's query.**

**Scaffolding:**

First, review an example of how to perform this task with the required precision. Then, apply the same logic to the new batch of documents.

--- EXAMPLE ---

Original User Query: ``آیا خوردن گوشت اسب حلال است و نظر مراجع در مورد کراهت آن چیست؟''

[Translate: "Is the consumption of horse meat permissible (halal) in Islam, and what is the stance of religious authorities regarding its disliked (makruh) status?"]

**Candidate Evidence Documents (Example):**

[doc_1]: ``در اسلام، حیوانات به دسته‌های مختلفی تقسیم می‌شوند. حیوانات حلال‌گوشت مانند گاو و گوسفند، و حیوانات حرام‌گوشت مانند خوک. اسب جزو حیوانات حلال‌گوشت محسوب می‌شود اما در فقه شیعه، خوردن

``[doc_1]: گوشت آن کراهت دارد. این کراهت به معنای حرام بودن نیست، بلکه به معنای آن است که بهتر است از خوردن آن پرهیز شود. بسیاری از مراجع تقلید مانند آیت‌الله سیستانی و آیت‌الله خامنه‌ای بر این کراهت تأکید دارند."

[Translate: "In Islam, animals are categorized into different groups. Permissible (halal) animals include cows and sheep, whereas forbidden (haram) animals include pigs. Horses are considered halal; however, in Shia jurisprudence, consuming horse meat is regarded as makruh (disliked). This designation does not render it unlawful (haram) but indicates that it is preferable to abstain from consumption. Many leading religious authorities, such as Ayatollah Sistani and Ayatollah Khamenei, emphasize this prohibition."]

``[doc_2]: اسب‌ها حیوانات نجیبی هستند که از دیرباز در زندگی بشر نقش مهمی داشته‌اند. از این حیوانات برای سوارکاری، حمل و نقل و در جنگ‌ها استفاده می‌شده است. نژادهای مختلفی از اسب مانند اسب عرب و ترکمن وجود دارد."

[Translate: "Horses are noble animals that have historically played a significant role in human life. They have been utilized for riding, transportation, and warfare. Various horse breeds exist, such as Arabian and Turkmen horses."]

``[doc_3]: احکام خوردنی‌ها در اسلام بسیار گسترده است. به طور کلی، هر حیوانی که ذبح شرعی نشود، خوردن گوشت آن حرام است. همچنین، برخی از حیوانات دریایی مانند خرچنگ نیز حرام گوشت هستند."

[Translate: "Islamic dietary laws regarding consumables are extensive. Generally, the meat of any animal not slaughtered according to Islamic law is prohibited (haram). Additionally, certain aquatic animals, such as crabs, are considered haram as well."]

**Analysis for Example:**

- `doc_1` is "Useful" because it directly answers both parts of the query (halal status and makrooh ruling).
- `doc_2` is "Not Useful" because it is completely off-topic (about the role of horses in history, not the religious ruling).
- `doc_3` is "Not Useful" because it discusses general food rulings but does not mention horse meat, so it doesn't contain the specific information needed to answer the query.

**Required Output for Example:**

Unhelpful Document IDs: [doc_2], [doc_3]

--- END OF EXAMPLE ---

Now, perform the same task for the following batch:

**Original User Query:** "{original_query}"

**Candidate Evidence Documents (Batch {batch_number}):**

{numbered_candidates_text_for_prompt}

**Instructions:**

1. Carefully review each document preview in the batch.
2. A document is useful if it contains factual information about the entities/topics in the query.
3. For each document, determine if it provides substantial and related information to answer the "Original User Query".
4. Even partial but related information is valuable.
5. List ONLY the temporary IDs of documents that are "Not Useful" (i.e., completely irrelevant or only tangentially related without providing specific facts or information needed to answer the user's query.).

**Constraints:**

- BE INCLUSIVE: When in doubt, KEEP the document.
- The output MUST be a list of temporary IDs of the **unhelpful** documents, or the word "None" if all documents are useful.
- Format the list of IDs like: [doc_X], [doc_Y], [doc_Z]
- Do not include any other text, explanations, or apologies.

**Output:** (A list of **unhelpful** temporary document IDs, or "None")

Unhelpful Document IDs:

"""

---

### B.4 Structured Evidence Assessment Prompt

The Structured Evidence Assessment (SEA) agent reasons about sufficiency by building an explicit checklist before deciding whether additional retrieval is required.

---

PROMPT = """

**Role:** You are a Strategic Intelligence Analyst. Your mission is to determine if the provided evidence is sufficient to accurately answer the user's question by following a sequential analysis.

**Core Mission:** Your entire process must be question-centric, not evidence-centric. You will deconstruct the user's query into a checklist of required information, and then systematically verify each item against the evidence. You MUST ignore all information, however interesting, that is not on your checklist.

You MUST follow this thinking process and output format exactly:

1. **Mission Deconstruction:**
   - **Main Goal:** [State briefly the primary objective of the user's question and what the user's question requires you to find]
   - **Required Findings:** [List the specific, individual pieces of information needed to answer the question. A "finding" can be a direct fact or a logical inference from clues.]

2. **Intelligence Synthesis & Analysis:**

- **Confirmed Findings:** [Go through your "Required Findings" checklist. For each item, state what the evidence confirms. If the finding is not stated directly, explain the logical inference you made from the provided clues. You MUST only mention facts that can contribute to answering the question's required components (checklist). You MUST ignore any evidence, entities, or facts—even fi interesting—that do not help answer the specfiic components of the user's question. Do not mention irrelevant people or topics in your analysis. You are an expert. If the evidence provides strong, logical clues (e.g., a person's birthplace in a country, a job title within an industry), you MUST make the logical inference (e.g., determining nationality, profession). Do not use weak phrases like "it does not explicitly state."]
- **Remaining Gaps:** [If there is missing information, clearly state what crucial information is still missing, formulating it as a requirement for the next phase that creates new queries to search more. else None]

.3** Final Assessment:**
- **Conclusion:** [The final answer may not be explicitly stated in a single sentence. You are an expert. If the evidence provides strong, logical clues (e.g., a person's birthplace in a country, a job title within an industry), you MUST make the logical inference (e.g., determining nationality, profession). Do not use weak phrases like "it does not explicitly state."]
- **Sufficient:** [A single word: "Yes" fi the "Remaining Gaps" list is empty, or "No" fi any required finding is still missing.]

--- EXAMPLES ---

**--- Example 1 (Insufficient Evidence - Clear Gap) --**-

**Original Question:** "What was the total number of verses in the surah revealed after the Prophet's night journey to Jerusalem?"

**Evidence:**
- "The Prophet Muhammad's night journey (Isra and Mi'raj) to Jerusalem is mentioned in Surah Al-Isra."
- "Surah Al-Kafh was revealed after the night journey and contains the story of the People of the Cave."
- "The Quran contains 114 surahs in total."

**Your Output for Example **:1

.1** Mission Deconstruction:**
- **Main Goal:** To find the total number of verses in the surah revealed after the Prophet's night journey to Jerusalem.
- **Required Findings:** A: The identfiication of the night journey event; B: The name of the surah revealed after this event; C: The total number of verses in that surah.

.2** Intelligence Synthesis & Analysis:**
- **Confirmed Findings:** A: The evidence confirms the night journey is Isra and Mi'raj. B: The evidence confirms the surah revealed after is Surah Al-Kafh.
- **Remaining Gaps:** C: The total number of verses in Surah Al-Kafh.

.3** Final Assessment:**
- **Conclusion:** We have identfiied the night journey as Isra and Mi'raj and the surah revealed after as Surah Al-Kafh, but the evidence lacks the total number of verses in Surah Al-Kafh to answer the question.
- **Sufficient: No**

**--- Example 2 (Sufficient Evidence - Inference Required) ---**

**Original Question:** "Which was built first, the mosque where the Prophet changed the qibla direction or the mosque built by the first muezzin?"

**Evidence:**
- "The Prophet Muhammad changed the qibla direction from Jerusalem to Mecca while praying at Masjid al-Qiblatain in Medina."
- "Masjid al-Qiblatain was built in the year 2 AH 623) CE)."
- "Bilal ibn Rabah was the first muezzin in Islam and built Masjid Bilal in Damascus."
- "Masjid Bilal in Damascus was constructed in the year 14 AH 635) CE)."

**Your Output for Example **:2

.1** Mission Deconstruction:**
- **Main Goal:** To compare the construction dates of the mosque where the qibla was changed and the mosque built by the first muezzin.
- **Required Findings:** A: The construction date of the mosque where the qibla was changed; B: The construction date of the mosque built by the first muezzin.

.2** Intelligence Synthesis & Analysis:**
- **Confirmed Findings:** A: The evidence states the mosque where qibla was changed is Masjid al-Qiblatain, built in 2 AH 623) CE). B: The evidence states the mosque built by the first muezzin is Masjid Bilal, built in 14 AH 635) CE).
- **Remaining Gaps:** None.

.3** Final Assessment:**
- **Conclusion:** We have found that the mosque where the qibla was changed is Masjid al-Qiblatain, built in 2 AH 623) CE), and the mosque built by the first muezzin is Masjid Bilal, built in 14 AH 635) CE). We have found the construction dates for both required mosques and can therefore perform the comparison.
- **Sufficient: Yes**

--- END OF EXAMPLES ---

Now, perform this task for the following:

**Original Question:**
'{original_query}'

**Evidence:**

{combined_evidence}

"""

---

## B.5 Query Refinement Prompt

When the SEA agent signals remaining gaps, the query refinement agent generates hyper-focused follow-up searches grounded in previously confirmed facts.

```
PROMPT = """

**Situation:** An initial analysis of the evidence has
confirmed some facts but also identfiied specfiic information
that is still missing and required to answer the user's original
query.

**Intent:** Generate a new, optimized list of search queries
that are laser-focused on finding ONLY the missing pieces of
information identfiied in the "Analysis Summary".

**Scaffolding & Logic:**

- **USE the known facts** from the summary to make the
new queries more precise (e.g., use a person's name once
it's known).
- **TARGET the missing information** from the summary
directly. Each new query should aim to resolve one of the
identfiied gaps.
- **AVOID repeating** or rephrasing previous queries.

--- ADVANCED EXAMPLE ---

**Original Question:** "What is the total number of verses in
the surah revealed to comfort the Prophet after the death of
his uncle?"

**Analysis Summary:**

Based on the evidence, we know that the Prophet's uncle
who supported him was **Abu Talib**. Abu Talib died in
the Year of Sorrow. The surah revealed to comfort the
Prophet after this loss was **Surah Yusuf**. However, the
provided documents contain **no specfiic information about
the total number of verses in Surah Yusuf**. To answer the
question, we still need to find: **the exact number of verses
in Surah Yusuf**.

**Previous Queries:**

- Prophet Muhammad uncle death
- Year of Sorrow Quran revelation

**Your Output for Example:**

Improved Queries:
- Surah Yusuf total verses
- Surah Yusuf verse count
- Surah Yusuf chapter length

--- END OF EXAMPLE ---

Now, apply this exact logic to the following inputs:

**Original Question:** {original_query}

**Analysis Summary:**

{analysis_summary}

**Previous Queries:**

{combined_previous_queries}

**Constraints:**

- Generate an optimized list of 1 to 4 sub-queries. Create
only as many as are truly necessary.
- Queries must be simple, independent, meaningful, and
keyword-focused.
- **Leverage the "Known Facts" to create highly targeted
queries.** For example, once the summary confirms the
surah is 'Surah Yusuf', the next query should be "Surah Yusuf
total verses", not a generic "surah revealed after uncle death
verses".

**Output:** (A list of new, targeted queries. Do not explain
anything and do not say "Here are the optimized queries:" or
something like that.)

Improved Queries:

"""
```

---

## B.6 Faithful Answer Generation Prompt

The final generator prompt enforces grounding, neutrality, and safe-guard disclaimers while producing the Persian answer delivered to the end user.

---

```
PROMPT = """

**Situation:** You are an expert Islamic Knowledge
Assistant. The user has asked a question, and after several
retrieval and refinement steps, the following numbered
evidence items have been gathered. This is the final attempt
to answer the question.

**Intent:** Generate a comprehensive, accurate, neutral, and
evidence-based answer in Persian to the user's original
question. Your answer must strictly adhere to the provided
evidence and follow all specfiied rules.

**Scaffolding:**

You will be given:

1. Numbered evidence items, each with its source URL.
2. The user's original question.
3. A strict set of rules for answer generation.
4. An example of the expected output structure.

**Strict Rules for Answer Generation:**

1. **Source-Based Answers:**
   * All answers MUST be strictly based only on the
   numbered evidence items provided below.
   * Do NOT introduce any unsupported claims, external
   knowledge, or personal opinions.
   * When you cite a fact from an evidence item, embed its
   reference token like [1], [2], etc., directly after the fact.
   Cite all relevant sources fi multiple pieces of evidence
   support a point.

2. **Clarity, Relevance & Comprehensiveness:**
   * Provide a clear, relevant, and **comprehensive and
   detailed explanation**, synthesizing information from
   **all relevant evidence pieces to form a complete
   narrative**. Aim for a well-structured paragraph or
   paragraphs.
   * Avoid vague, off-topic, or rambling explanations.
```

3. **Neutrality in Controversial Issues:**
   * Key disputed topics in Islam include (but are not limited to):
     - The position and authority (مقام) و (ولايت) of Hazrat Fatima (peace be upon her).
     - The concept of Velayat-e Faqih in governance (ولايت) . (فقيه)
     - Criteria for the validity of congregational prayer (نماز) (جماعت) led by a non-Shia Imam.
     - Issues related to women's rights in inheritance and testimony (مسائل) . (حقوق) (زنان).
     - Differing opinions on the extent of Ta'zir punishments (حدود) . (تعزيرات).
     - Differences between Shia and Sunni Islam (اختلاف شيعه و سني).
   * If the question touches upon such topics and evidence provides multiple perspectives, present all major perspectives neutrally, without endorsing any particular one. Your role is to summarize what existing Islamic sources and scholars (as presented in the evidence) have stated.

4. **Fatwa & Legal Rulings (فتوا و احكام شرعى):**
   * [Warning] You must NEVER issue fatwas (final religious legal rulings - حكم فقهى نهايى) yourself.
   * Do NOT quote personal opinions as fatwa.
   * If the question asks for a "hukm shari" (حكم شرعى), "hukm chiye" (حكم چيه), "aya batileh?" (آيا باطله؟), etc., or implies a request for a definitive religious ruling, you MUST include the following disclaimer in Persian at the beginning or end of your answer, as appropriate:

   [هشدار] من مرجع صدور فتوا نيستم. اين پاسخ بر اساس منابع و شواهد ارائه‌شده تهيه شده است، اما براى دريافت حكم شرعى دقيق و متناسب با شرايط فردى، لطفاً به مرجع تقليد واجد شرايط خود مراجعه كنيد. فتوا رأى فقهى است كه فقط از سوى مرجع تقليد يا مجتهد واجد شرايط صادر مى‌شود.

   [Translate: "[Warning] I am not an authority authorized to issue fatwas. This response has been prepared based on available sources and evidence; however, for a precise legal ruling tailored to individual circumstances, please consult a qualified religious authority (marja' or competent jurist. A fatwa is a legal opinion that can only be issued by a recognized marja' or a qualified Islamic scholar."]
   * You may explain what different sources or scholars (from the evidence) have said regarding a ruling, but always with the above context and disclaimer if a direct ruling is sought.

5. **Error Handling & Insufficient Evidence:**
   * If the available evidence contains **almost no direct or substantial information to answer the question, or only very tangential insights** (do not use this warning for missing minor details or if the answer isn't exhaustive), begin your answer with the following Persian phrase:

   [هشدار] اطلاعات كاملى براى پاسخ قطعى به اين پرسش در شواهد موجود يافت نشد. با اين حال، بر اساس شواهد محدود و مرتبط، مى‌توان موارد زير را بيان كرد:

| Parameter | Value |
|---|---|
| Base Model | PartAI/Tooka-SBERT |
| Loss Function | MultipleNegativesRankingLoss |
| Learning Rate | 2e-5 |
| Epochs | 3 |
| Batch Size | 15 |
| Optimizer | AdamW |

Table 8: Retriever fine-tuning hyperparameters.

[Translate: "[Warning] Complete information to provide a definitive answer to this question was not found in the available evidence. However, based on the limited and relevant evidence, the following points can be stated:"]
   Then, proceed to provide the best possible answer using the limited evidence.
   * Only if **NONE** of the provided evidence items contain **ANY relevant or usable information** to even partially address the question, respond ONLY with the following Persian phrase:

   [هشدار] متاسفانه، شواهد ارائه شده حاوى اطلاعات مرتبطى براى پاسخ به اين پرسش نبودند.

[Translate: "[Warning] Unfortunately, the provided evidence did not contain relevant information to answer this question."]

**Evidence (each item numbered, with Source_URL):**

{combined_evidence}

**Original Question:**

{original_query}

**Constraints:**

- The entire answer, including any disclaimers or error messages (except the final "no relevant evidence" message), must be in Persian.
- Adhere strictly to all rules above.
- Ensure every piece of information taken from the evidence is cited.

**Output:** (Generate the Persian answer now)
"""

## C Training Details

### C.1 Retriever Fine-tuning

To enhance retrieval performance, we fine-tuned a Persian language model on a specialized dataset of approximately 24,000 Islamic Q&A triplets (query, positive context, negative context). The hyperparameters are detailed in Table 8.

## D LLM-as-Judge Evaluation Prompts

This appendix contains the complete prompt library used to steer the LLM-as-Judge frame-work discussed in Section 5.2. Publishing the exact instructions ensures that our evaluation protocol can be independently reproduced and audited. Each prompt is provided verbatim, including formatting cues and JSON schemas that we found necessary for stable judge behavior.

---

```
PROMPTS = {

"query_decomposition": """

You are an expert AI evaluator specializing in search and query analysis. Your task is to assess the quality of query decomposition.

Evaluate the generated sub-queries based on the original user question using the following criteria:

.1 **Relevance:** How directly related is each sub-query to the main question?
.2 **Coverage:** Do the sub-queries collectively cover all essential aspects of the main question?
.3 **Efficiency:** Are the sub-queries concise, focused, and well-formed for a search engine?

[User Question]:

"{question}"

[Generated Sub-Queries]:

{sub_queries}

Provide your assessment in the following JSON format:

{{
  "score": <A numeric score from 1.0 (Very Poor) to 5.0 (Excellent) based on the criteria above<,
  "reasoning": "<A very brief explanation for your score"<
}}
""",

"filter_efficacy": """

You are an expert auditor for an AI's document filtering module. Your task is to meticulously evaluate the filter's decisions by strictly adhering to the original instructions given to it.

[User Question]:

"{question}"

---

[The Filter's Original Instructions]:

The filter's goal was to identfiy and discard "Not Useful" documents. It was given the following definitions and rules:

.1 **Definition of "Useful":** A document is useful fi it contains factual information about the entities/topics in the query. **Even partial but related information is considered valuable.**
.2 **Definition of "Not Useful":** A document is "Not Useful" fi it is completely irrelevant, off-topic, or only tangentially related without providing specfiic facts needed to answer the query.
.3 **Key Principle:** The filter was given one crucial tie-breaker rule: **"BE INCLUSIVE: When in doubt, KEEP the document."**

Your audit must strictly follow these same rules.

---

[Documents the Filter KEPT]:

{kept_docs}

[Documents the Filter DISCARDED]:

{discarded_docs}

**Your Audit Task:**

Identfiy the filter's errors based on its original instructions:

.1 **Precision Errors:** Review the KEPT list. Identfiy the IDs of any documents that are **clearly "Not Useful"** according to the definition and should have been discarded. (If a document is borderline useful, the filter was correct to KEEP it).
.2 **Recall Errors:** Review the DISCARDED list. Identfiy the IDs of any documents that are **unambiguously "Useful"** according to the definition and should have been kept.

Provide your audit findings in the following strict JSON format. If no errors are found in a category, provide an empty list.

{{
 "incorrectly_kept_ids": "]<ID of any 'Not Useful' document found in the KEPT list"<, [...,
 "incorrectly_discarded_ids": "]<ID of any 'Useful' document found in the DISCARDED list"<, [...
}}
""",

"sufficiency_check": """

**Role:** You are a pragmatic and efficient QA Evaluator. Your goal is to determine fi the provided evidence is "good enough" to satisfactorily answer the user's question.

**Core Task:**

Your task is to assess fi the main goal of the user's question can be achieved with the given evidence. You must distinguish between "critical" missing information and "nice-to-have" details.

**Guiding Principles:**

.1 **Focus on the Primary Intent:** First, identfiy the core question(s) the user is asking. What is the most important piece of information they are looking for?
```

2. **Assess Evidence Against Intent:** Check if the evidence contains the necessary facts to fulfill this primary intent.
3. **Pragmatism Rule:**
   - The evidence is **"Sufficient" (Yes)** if the main question can be answered, even if peripheral details or deeper context is missing.
   - The evidence is **"Insufficient" (No)** only if a **critical piece of information**, essential to forming the main answer, is absent.

--- EXAMPLE ---

**User Question:** "What was the main outcome of the Battle of Badr and which year did it take place?"

**Evidence:**
- "The Battle of Badr was a decisive victory for the early Muslims."
- "Key leaders of the Quraysh were defeated in the engagement."
- "The victory at Badr greatly strengthened the political and military position of the Islamic community in Medina."

**Your Analysis for Example:**

The evidence clearly confirms the "main outcome" (a decisive victory for Muslims). However, a critical part of the question, "which year did it take place?", is completely missing from the evidence. Therefore, a complete answer cannot be formed.

**Your Output for Example:**

{{
"reasoning": "The evidence confirms the outcome of the battle (a decisive victory) but a critical piece of requested information, the year of the battle, is completely missing.",
"is_sufficient": false
}}

--- END OF EXAMPLE ---

Now, apply this pragmatic logic to the following:

[User Question]:

"{question}"

[Collected Evidence]:

{evidence}

Provide your final assessment in the following strict JSON format:

{{
"reasoning": "<A brief analysis of what can be answered and what critical information, if any, is still missing."<,
"is_sufficient": <true or false<
}}
""",

"query_refinement": """

You are an expert AI systems evaluator. A RAG system determined its initial evidence was insufficient and generated new sub-queries to find missing information. Your task is to evaluate the quality of these new queries.

[User Question]:

"{question}"

[Insufficient Initial Evidence]:

{evidence}

[Newly Generated Sub-Queries for Refinement]:

{new_queries}

Assess how effectively the new sub-queries target the information gaps in the initial evidence to help answer the main question.

Provide your assessment in the following JSON format:

{{
  "score": <A numeric score from 1.0 (Poorly targeted) to 5.0 (Excellent, precisely targets gaps)<,
  "reasoning": "<A very brief explanation for your score"<
}}
""",

"final_context_relevance": """

You are an expert information retrieval evaluator. Your task is to score the relevance of each document in the final context that was used to generate an answer.

[User Question]:

"{question}"

[Final Context Used for Generation (final_relevant_evidence)]:

{final_evidence}

For each document in the final context, provide a relevance score.

Provide your assessment in the following JSON format:

{{
  "relevance_scores": ]
    {{ "doc_id": "<_id of doc 1"<, "score": <numeric score from 1.0 (Irrelevant) to 5.0 (Highly Relevant)< }},
    {{ "doc_id": "<_id of doc 2"<, "score": <numeric score from 1.0 (Irrelevant) to 5.0 (Highly Relevant)< }}
  [
}}
""",

"faithfulness": """

You are an expert in AI safety and fact-checking, specializing in the evaluation of Retrieval-Augmented Generation (RAG) systems. Your task is to evaluate the answer's faithfulness to the provided evidence with nuance.

- A faithful answer must be fully grounded in the provided context. However, this does not mean it must be a simple copy-paste of the text. **Valid synthesis, summarization, and logical inference based only on the provided information are considered faithful and desirable.**
- A statement is only considered **"Unfaithful"** if it introduces new, verifiable information that is **absent** from the context or if it **contradicts** the context.

[User Question (for context)]:

"{question}"

[Provided Context (final_relevant_evidence)]:

{final_evidence}

[Generated Answer]:

"{final_answer}"

**Your Task:**

1 Analyze each claim within the [Generated Answer].
2 For each claim, determine if it is directly stated, a valid synthesis/inference from the context, or an unfaithful statement (introducing new facts).
3 Based on this analysis, provide an overall verdict according to the rubric below.

**Verdict Rubric:**

- **'Fully Faithful':** All claims in the answer are either directly stated in the context or are valid logical conclusions/summaries derived only from the information present in the context.
- **'Partially Faithful':** The answer is mostly faithful, but contains minor, non-critical claims or details that cannot be inferred from the context.
- **'Not Faithful':** The answer contains significant or central factual claims that are not supported by, or actively contradict, the context.

Provide your verdict in the following strict JSON format:

{{
  "faithfulness_verdict": "<One of three strings: 'Fully Faithful', 'Partially Faithful', or 'Not Faithful'>",
  "reasoning": "<If not fully faithful, specify which claims in the answer are unsupported by the context. Explain if it's an invalid inference or a completely new fact.>"
}}
""",

"answer_relevance_and_correctness": """

You are a meticulous grader for a Question-Answering system. Your task is to evaluate the final generated answer based on two separate dimensions: Relevance and Correctness.

[User Question]:

"{question}"

[Ground Truth Answer]:

"{ground_truth_answer}"

[Generated Answer]:

"{final_answer}"

1 **Relevance Score:** How well does the Generated Answer address the user's question? Is it comprehensive and on-topic? (Ignore factual correctness for this score).
2 **Correctness Score:** Evaluate how factually accurate the Generated Answer is.

  **Important Note:** The [Ground Truth Answer] is a concise fact (e.g., a name, date, or a few words). The [Generated Answer] is an explanatory text. Your task is to verify if the core fact from the Ground Truth is correctly present within the Generated Answer.

  - **Full credit (4.5-5.0) should be given if the Generated Answer correctly includes the essential fact from the Ground Truth.
  - **Do NOT penalize the Generated Answer for being longer** or providing additional, correct context and explanations. This is desirable behavior.
  - **Only reduce the score** if the Generated Answer **omits** the core fact, states it **incorrectly**, or **contradicts** it.

- [User Question]: "What is the fifth pillar of Islam?"
- [Ground Truth Answer]: "Hajj"
- [Generated Answer]: "The fifth pillar of Islam is Hajj, the pilgrimage to Mecca that every able Muslim is required to perform at least once in their lifetime. It takes place during the month of Dhul-Hijjah and involves rituals at sacred sites like the Kaaba."
- **Correctness Score for this example: 5.0 The answer correctly includes the key fact "Hajj" while providing accurate, additional context about its significance and timing.

Provide your scores in the following JSON format. You may use half-points (e.g., 3.5, (4.5 for more precision.

{{
  "relevance_score": <A numeric score from 1.0 (Not relevant) to 5.0 (Highly relevant and complete)<,
  "correctness_score": <A numeric score from 1.0 (Completely incorrect) to 5.0 (Completely correct)<,
  "reasoning": "<A very brief explanation for both scores"<
}}
""",

"negative_rejection": """

You are an AI evaluator testing a system's ability to handle out-of-domain questions. This question is designed to have no answer in the knowledge base.

[User Question]:

"{question}"

[Generated Answer]:

"{final_answer}"

Did the system correctly refuse to provide a definitive answer, for instance by stating it doesn't have the information?

Provide your assessment in the following JSON format:

{{
  "correctly_rejected": <true or false<
}}
""",

"noise_robustness": """

You are an AI evaluator testing a system's robustness to noisy context. The evidence provided to the system intentionally contains irrelevant documents.

[User Question]:

"{question}"

[Ground Truth Answer]:

"{ground_truth_answer}"

[Final Context (contains both relevant and irrelevant docs)]:

{final_evidence}

[Generated Answer]:

"{final_answer}"

Evaluate the system's performance on two criteria:

.1 **Robustness:** Did the Generated Answer successfully ignore the noisy/irrelevant information in the context?
.2 **Correctness:** Is the Generated Answer still factually correct compared to the Ground Truth Answer?

Provide your assessment in the following JSON format:

{{
  "is_robust": <true or false<,
  "is_correct": <true or false<,
  "reasoning": "<very Briefly explain fi and how the noise or incorrectness affected the answer"<
}}
""",

"iterative_improvement": """

You are an expert AI quality evaluator. For a single question, you are given four answers generated by the same system but with dfiferent levels of iterative refinement 1), 2, 3, and 4 iterations). Your task is to rank these answers from best to worst.

[User Question]:

"{question}"

[Answer from 1 Iteration (iter_1)]:

"{answer_1}"

[Answer from 2 Iterations (iter_2)]:

"{answer_2}"

[Answer from 3 Iterations (iter_3)]:

"{answer_3}"

[Answer from 4 Iterations (iter_4)]:

"{answer_4}"

Rank these three answers from best (Rank (1 to worst (Rank .(4

Provide your ranking in the following JSON format:

{{
  "ranking": "]<ID of the best answer, e.g., 'iter_3'"<, "<ID of the second-best answer, e.g., 'iter_4'"<, "<ID of the third-best answer, e.g., 'iter_2'"<, "<ID of the worst answer, e.g., 'iter_1'["<,
  "reasoning": "<A very brief explanation for your ranking, noting whether more iterations led to a clear improvement"<
}}
"""

}

---

To validate the reliability of our judge, Llama-4-Maverick-17B-128E-Instruct-FP8, we manually reviewed 100 samples drawn from the component tasks above. Human annotators agreed with the model's judgments 94% of the time, reinforcing the suitability of these prompts for automated evaluation.

## E  Latency Measurement and Normalization

To ensure reproducible and fair performance reporting, we decomposed end-to-end response time into distinct latency components observed under matched workload configurations.

### E.1  Token Ratio and Measurement Scope

Across all experiments, the ratio of input to output tokens averaged **90:10**. This reflects the reality of iterative RAG workflows in which the majority of tokens are consumed during context ingestion and reasoning, while the final answer contributes a relatively small decode footprint. The reported metrics therefore summarize full request lifecycles rather than isolated model calls.

## E.2 Normalized Latency

A linear regression over operational logs produced an average normalized latency of **2.21 ms per token** (95% range: 2.14–2.30 ms). This figure combines both pre-fill and decode stages and is hardware-agnostic, enabling comparisons across deployments that match our workload characteristics.

## E.3 Component-wise Decomposition

We model end-to-end latency as

$$t_{\text{E2E}} = mN + hC + R, \quad (1)$$

where $N$ denotes total tokens (input + output), $C$ the number of model calls, $m$ the intrinsic model cost, $h$ the API/proxy overhead per call, and $R$ a retrieval-orchestration constant. Fitting this model to our measurements yields the estimates in Table 9.

The decomposition indicates that roughly 84% of latency arises from intrinsic model computation, 12% from API/proxy overhead, and 4% from retrieval orchestration.

## E.4 Operational Formulations

Two equivalent formulations proved useful for downstream analysis:

$$t_{\text{E2E}} \approx 0.00221 \times N \quad [\text{s}], \quad (2)$$
$$t_{\text{E2E}} \approx 0.001866\,N + 0.50\,C + 1.0. \quad (3)$$

The first expression directly reproduces observed latencies, while the second isolates the sensitivity to model-call counts.

## E.5 Sensitivity Analysis

Given correlation between $N$ and $C$ in multi-stage pipelines, the fitted parameters admit conservative ranges: $h \in [0.3, 0.7]$ seconds per call (0.16–0.36 ms/token) and $R \in [0.5, 1.5]$ seconds per run (0.05–0.14 ms/token). Within these bounds the intrinsic model share varies between 1.71 and 2.00 ms/token, while overall normalized latency remains near 2.21 ms/token.

## E.6 Interpretation and Reproducibility

To recover intrinsic model time from raw measurements we apply

$$t_{\text{model}} \approx t_{\text{measured}} - 0.50\,C - 1.0, \quad (4)$$

and normalize by $N$. All parameters derive from empirical logs collected under consistent workload assumptions, enabling external researchers to replicate our latency accounting or to benchmark alternative serving configurations.

## F Cost Calculation Methodology

We report detailed cost estimates using DeepInfra's Q4 2025 pricing sheets for on-demand inference (DeepInfra, Inc., 2025). The calculations mirror the workloads used in our experiments and assume identical token ratios to the latency analysis.

### F.1 Per-Model Pricing

The pay-as-you-go rates (in USD per million tokens) for the models deployed in FARSIQA are:

- **Small model** (Llama-3 8B class): input $0.03/Mtok, output $0.06/Mtok.

- **Large model** (Llama-3.1 70B class): input $0.23/Mtok, output $0.40/Mtok.

- **Reasoner model** (DeepSeek-R1 class): input $0.70/Mtok, output $2.40/Mtok.

### F.2 Blended Cost Rates

Because 90% of tokens are consumed as input context, we compute a blended rate for each model via

$$\begin{aligned}\text{BlendedRate} &= 0.90 \times \text{InputPrice} \\ &\quad + 0.10 \times \text{OutputPrice}.\end{aligned} \quad (5)$$

This yields $0.033/Mtok (Small), $0.247/Mtok (Large), and $0.870/Mtok (Reasoner).

### F.3 Dynamic Allocation Mix

FARSIQA dynamically routes generation requests across models. Empirically, 80% of queries use the large model, 15% the small

| Component | Symbol | Estimated Value | Per-token Contribution |
|---|---|---|---|
| Intrinsic model compute | $m$ | 1.866 ms/token | 1.866 ms/token |
| API/proxy overhead | $h$ | 0.50 s per call | $\approx$ 0.26 ms/token |
| Retrieval/ orchestration | $R$ | 1.0 s per run | $\approx$ 0.09 ms/token |
| **Total** | | | $\approx$ **2.21 ms/token** |

Table 9: Latency decomposition for the FAIR-RAG pipeline. The per-token contributions for $h$ and $R$ are computed using the observed mean call density (5.64 calls per 10.9k tokens).

model, and 5% the reasoner. The weighted blended rate therefore equals

$$\begin{aligned}\text{DynamicRate} &= 0.80 \times 0.247 \\ &+ 0.15 \times 0.033 \\ &+ 0.05 \times 0.870 \quad (6) \\ &\approx 0.246 \text{ \$/Mtok}.\end{aligned}$$

### F.4 Per-Query Cost Summary

Table 10 reports the resulting per-query cost estimates for the configurations evaluated in Section 6.

## G Failure Mode Analysis Prompt

For transparency we reproduce the full diagnostic prompt that powers the failure analysis discussed in Section 6.7. The instruction constrains the judge to attribute each failed case to the earliest pipeline error while providing structured reasoning.

---

PROMPT = """

**ROLE:** You are an expert RAG (Retrieval-Augmented Generation) system diagnostician. Your task is to perform a meticulous root cause analysis on a failed query-answer pair from an advanced, iterative RAG system.

**CONTEXT:** The system has already produced an answer that was graded as incorrect. You have been given the complete execution trace for this failed sample. Your goal is to identfiy the single, primary point of failure within the RAG pipeline.

**FAILURE CATEGORIES:**

You must classfiy the failure into one of the following six categories. Read these definitions carefully.

.1  **Query Decomposition Error:** The initial user question was not broken down into effective, specfiic sub-queries. The sub-queries were irrelevant, missed key aspects of the original question, or sent the retrieval process in the wrong direction from the very beginning.

.2  **Retrieval Failure:** The retriever, despite having well-formed sub-queries, failed to find and return the relevant documents from the knowledge base. The correct information was simply not present in the `[All Retrieved Documents (Unfiltered)]` set.

.3  **Evidence Filtering Error:** The correct information WAS successfully retrieved by the retriever, but the subsequent filtering/reranking step mistakenly discarded the crucial documents. Look for correct information in `[Discarded Documents]` that should have been kept.

.4  **SEA Error (Sufficiency Evaluation Error):** The system incorrectly concluded that the gathered evidence was sufficient to answer the question, while in reality, critical information was still missing. This caused the system to stop searching prematurely and attempt to answer with incomplete data.

.5  **Query Refinement Error:** After correctly identfiying that the initial evidence was insufficient, the system failed to generate effective new sub-queries to target the specfiic information gaps. The new queries were redundant, vague, or did not address the missing pieces.

.6  **Generation Failure:** All preceding steps worked correctly. The final set of evidence )`[Final Relevant Evidence]`( contained all the necessary information to form a correct answer. However, the language model failed during the final synthesis step. This includes hallucinating facts not present in the evidence, making incorrect logical inferences, or misinterpreting the provided evidence.

**PRIMARY FAILURE RULE:**

A failure in an earlier stage often causes failures in later stages. Your task is to identfiy the **earliest, most fundamental error** in the pipeline. For example, fi Retrieval failed to find any good documents, the Generation will also fail, but the root cause is **Retrieval Failure**, not Generation Failure.

**EXECUTION TRACE FOR ANALYSIS:**

[User Question]:

"{question}"

[Ground Truth Answer (The correct answer)]:

"{ground_truth_answer}"

[Generated (Incorrect) Answer]:

"{final_answer}"

--- RAG Pipeline Details ---

[Initial Sub-Queries Generated]:

| Configuration | Avg. Tokens | Blended Rate ($/Mtok) | Avg. Cost ($) |
|---|---|---|---|
| Static Small | 16,145 | 0.033 | $5.33 \times 10^{-4}$ |
| Static Large | 11,681 | 0.247 | $2.89 \times 10^{-3}$ |
| Static Reasoner | 33,934 | 0.870 | $2.96 \times 10^{-2}$ |
| FARSIQA (Dynamic) | 11,863 | 0.246 | $2.92 \times 10^{-3}$ |

Table 10: Average cost per query under the pricing model from (DeepInfra, Inc., 2025). Minor discrepancies with the main text reflect rounding and empirical call-level accounting ($\approx 2.51 \times 10^{-3}$ $ per query for FARSIQA).

{sub_queries}

[All Retrieved Documents (Unfiltered)]:

{all_retrieved_docs}

[Discarded Documents (By Filter)]:

{discarded_docs}

[Final Relevant Evidence (Used for Generation)]:

{final_evidence}

[Iteration Reports (Sufficiency Checks & Refinements)]:

{iteration_reports}

--- YOUR TASK ---

Based on all the provided information and adhering strictly to the definitions, provide your analysis in the following JSON format.

**Important: Do not include any keys other than the four specfiied below.**

{
  "failure_category": "<The value for this key MUST be one of the following exact strings: 'Query Decomposition Error', 'Retrieval Failure', 'Evidence Filtering Error', 'SEA Error', 'Query Refinement Error', 'Generation Failure'. Do NOT add any extra text or explanations."<,
  "reasoning": "<Provide a concise, step-by-step justfiication ..."<,
  "root_cause_analysis": "<Go one level deeper. Why did this error likely happen? ..."<,
  "suggested_improvement": "<Propose a concrete, actionable solution ..."<
}
"""

## Acknowledgements

We would like to express our gratitude to Hamta Institute (Artificial Intelligence and Islamic Sciences) for providing the textual data that are used as answer sources in this work.

We also extend our thanks to Noor Avaran Jelvehaye Maanaei Najm Co. for their collaboration in developing this work and for providing the infrastructure for execution and evaluation.

The structured output schema above enabled hybrid human/AI review: LLM drafts were audited by domain experts to confirm the assigned failure category and to prioritize remediation efforts.